\documentclass{article}

\usepackage[final]{neurips_2021}

\usepackage[utf8]{inputenc} % allow utf-8 input
\usepackage[T1]{fontenc}    % use 8-bit T1 fonts
\usepackage{hyperref}       % hyperlinks
\usepackage{url}            % simple URL typesetting
\usepackage{booktabs}       % professional-quality tables
\usepackage{amsfonts}       % blackboard math symbols
\usepackage{nicefrac}       % compact symbols for 1/2, etc.
\usepackage{microtype}      % microtypography
\usepackage{xcolor}         % colors
\usepackage{multirow}
\usepackage{graphicx}
\usepackage{caption}
\usepackage{subcaption}
\usepackage{pgfplots}
\usepackage{pgfplotstable}
\usetikzlibrary[patterns]
\usepackage[inline]{enumitem}
\usepackage{amsmath}
\pgfplotsset{compat=1.11}
\usepackage{verbatim}
\usepackage{dirtree}
\usepackage{makecell}
\usepackage{algorithm}
\usepackage{algpseudocode}
\usepackage{algorithmicx}
\usepackage{multicol,multirow}
\usepackage[scientific-notation=true]{siunitx}

\definecolor{metagray}{RGB}{28,43,51}
\definecolor{metablue}{RGB}{0,75,185}
\definecolor{metaorange}{RGB}{250, 135, 25}
\definecolor{metadarkblue}{RGB}{0,30,110}

\newcommand{\GRN}{MMS-unlab} % Global Recordings network
\newcommand{\MMS}{MMS-lab} %New testament corpus
\newcommand{\MMSU}{MMS-lab-U} %New testament corpus
\newcommand{\MMSUboth}{MMS-lab-U+unlab} % both

\newcommand{\modelname}{MMS}
\newcommand{\modelnameb}[1]{\modelname{} {(#1B)}}

\newcommand{\numptlang}{1,406}          % Number of pretraining languages
\newcommand{\numgrnlang}{3,809}         % Number of lang in GRN dataset
\newcommand{\nummmslang}{1,107}         % Number of lang in MMS datast
\newcommand{\nummmsulang}{1,362}         % Number of lang in MMS-U datast
\newcommand{\numlidlang}{4,017}  % Number of lang for LID models

\newcommand{\numgrnhrs}{7.7K} % Number of hours in GRN dataset
\newcommand{\nummmshrs}{44.7K}  % Number of hours in MMS dataset
\newcommand{\nummmshrsrd}{45K}  % Number of hours in MMS dataset rounded
\newcommand{\nummmsuhrs}{55K} % Number of hours in MMS-U dataset

\DeclareMathOperator*{\argmax}{arg\,max} % thin space, limits underneath in displays

\newcommand{\ci}[1]{{\tiny $\pm$ #1}}

\newcommand{\Inp}{\mathcal{X}}
\newcommand{\Feat}{\mathcal{Z}}
\newcommand{\QFeat}{\mathcal{Q}}

\newcommand{\Context}{\mathcal{C}}

\newcommand{\Star}{\langle \ast \rangle}

\newcommand{\Alphabet}{\mathcal{A}}
\newcommand{\CollapseFunc}{\mathcal{B}}
\newcommand{\Blank}{\langle b \rangle}

\newcommand{\ze}{z}
\newcommand{\zq}{q}

\newcommand{\cc}{c}

\newcommand{\Enot}[1]{\num[exponent-product = \times]{#1}}

\newcommand{\shellcmd}[1]{\indent\texttt{\footnotesize #1}}

\newcommand{\insertDatasetLanguageComparison}{
\begin{figure}%
  \centering
  \begin{subfigure}[t]{0.5\linewidth}
    \centering
    \resizebox{\linewidth}{!}{

    \begin{tikzpicture}

      \begin{axis}[
          width=\linewidth,
          enlargelimits=0.2,
          xmode=log,
          ymode=log,
          xlabel={\# languages},
          ylabel={\# hours},
          xmin=5, xmax=4500,
          ymin=1000, ymax=450000,
          ytick={1000, 10000, 100000,1000000},
          yticklabels = {$10^3$, $10^4$, $10^5$, $10^6$},
          legend pos=north west,
          ymajorgrids=true,
          grid style=dashed,
          log ticks with fixed point,
        ]
        \addplot[
          scatter, only marks,
          scatter/classes={
              a={mark = o, black},
              c={mark = o, black},
              b={mark = *, metablue}
            },
          point meta=explicit symbolic,
          visualization depends on={value \thisrow{z} \as \Label},
          visualization depends on={\thisrow{yshift} \as \yShift},
          visualization depends on={\thisrow{xshift} \as \xShift},
          nodes near coords*={\Label},
          nodes near coords style={   
                yshift=\yShift,
                xshift=\xShift,
         },
        ]
        table[row sep=\\,col sep=&,meta=class]{
            x    &   y    & class & z & xshift & yshift\\
            8  &   49400   &  a & MLS  & 0  & 0\\
            25 &1400 & c & BABEL  & 16 & 0 \\
            16 & 1800 & a & VoxPopuli  & -7 & 2 \\
            98 &17127& a & CommonVoice  & 0 & 0\\
            102 &1200 & a & FLEURS  & 4 & -13\\
            8 &999 & a & M-AILABS  & -4 & -13\\
            699 & 13725 & a & CMU Wilderness  & 0 & -14 \\
            1107 & 44700& b & {\MMS}  & 0 & 0\\
          };
      \end{axis}
   \end{tikzpicture}
}
\label{fig:lab-speech-datasets}
\caption{Labeled Datasets.}
\end{subfigure}\hfill 
  \begin{subfigure}[t]{0.5\linewidth}
  \centering
    \resizebox{\linewidth}{!}{
    \begin{tikzpicture}
      \begin{axis}[
          width=\linewidth,
          enlargelimits=0.2,
          xmode=log,
          ymode=log,
          xlabel={\# languages},
          ylabel={\# hours},
          xmin=5, xmax=4500,
          ymin=1000, ymax=450000,
          ytick={1000, 10000, 100000,1000000},
          yticklabels = {$10^3$, $10^4$, $10^5$, $10^6$},
          legend pos=north west,
          ymajorgrids=true,
          grid style=dashed,
          log ticks with fixed point,
        ]
        \addplot[
          scatter, only marks,
          scatter/classes={
              a={mark = o, black},
              b={mark = *, metablue}
            },
          point meta=explicit symbolic,
          visualization depends on={value \thisrow{z} \as \Label},
          visualization depends on={\thisrow{yshift} \as \yShift},
          visualization depends on={\thisrow{xshift} \as \xShift},
          nodes near coords*={\Label},
          nodes near coords style={   
                yshift=\yShift,
                xshift=\xShift,
         },
        ]
        table[row sep=\\,col sep=&,meta=class]{
            x    &   y    & class & z  & xshift & yshift\\
            23 & 400000& a & VoxPopuli &0 & 0\\
            107 & 6200 & a & VoxLingua107 &0 & -15 \\
            40 & 17000 & a & Dhwani &0  & 0 \\
            3809 & 7700 & b & \GRN & -3 & 0 \\
            1362 & 55000 & b & \MMSU & -3 & 0 \\
          };
      \end{axis}
    \end{tikzpicture}
    
  }
  \label{fig:unlab-speech-datasets}
  \caption{Unlabeled Datasets.}
  \end{subfigure}
\caption{
\textbf{Dataset Overview.}
\MMS{}, \MMSU{} and \GRN{} compared to existing multilingual speech corpora in terms of the supported languages and dataset size. 
We compare to BABEL~\citep{gales2014babel}, CMU Wilderness~\citep{black2019cmu}, CommonVoice~\citep{ardila2019common}, Dhwani~\citep{javed2022towards}, FLEURS~\citep{conneau2022fleurs}, M-AILABS~\citep{mailabs}, MLS~\citep{pratap2020mls}, VoxLingua107~\citep{valk2020voxlingua}, and VoxPopuli~\citep{wang2021voxpopuli} .
}
\label{fig:datasets-comp}
\end{figure}
}

\newcommand{\insertGenderAnalysis}{
\definecolor{metagray}{RGB}{28,43,51}
\definecolor{metablue}{RGB}{0,75,185}
\definecolor{metadarkblue}{RGB}{0,30,110}
\begin{figure}
\centering
\begin{tikzpicture}
\begin{axis}[
ybar,
bar width=.15in,
width=.55*\textwidth,
height=.4\textwidth,
legend style={at={(0.95,0.98)},draw=none},
legend cell align={left},
legend columns=2,
nodes near coords align={vertical},
ylabel={CER},
ymin=0,
ymax=15,
symbolic x coords={Overall,Female,Male},
enlarge x limits=0.3,
xtick=data]
\addplot[fill=metablue] coordinates {
(Overall,12.37)
(Female,12.44)
(Male,12.26)
};
\addplot[fill=metagray, pattern= north east lines] coordinates {
(Overall,6.35)
(Female,6.49)
(Male,6.24)
};
\legend{\modelname{} data, FLEURS data}
\end{axis}
\end{tikzpicture}
\caption{
\textbf{Analysis of Gender Bias.}
We compare the character error rate (CER) of automatic speech recognition models trained on \MMS{} data and FLEURS data for male and female spakers.
Results are on the development sets of 27 languages of the FLEURS benchmark for which \MMS{} provides data and for which there are at least 50 samples for each gender.
}
\label{fig:rai-gender-analysis}
\end{figure}
}

\newcommand{\insertGRNHours}{
\begin{figure}[t]
  \centering
\includegraphics[width=\linewidth]{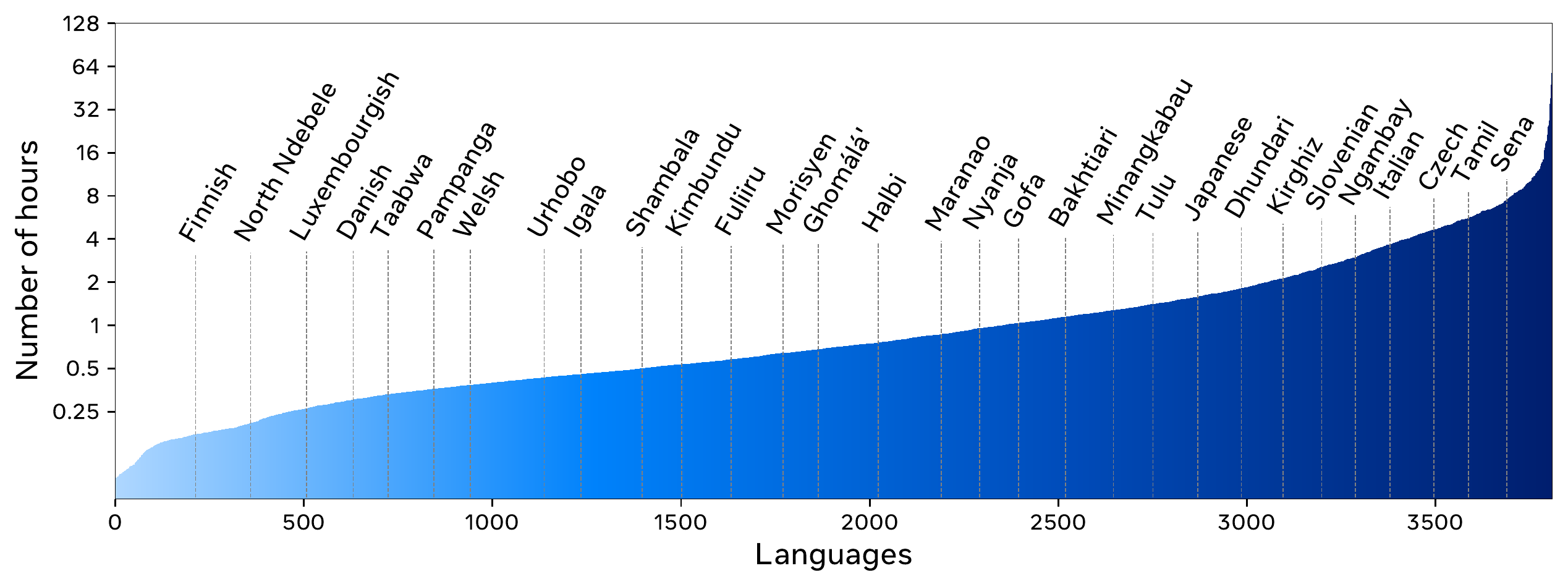}
\caption{
\textbf{\GRN{}: Amount of Speech Data across Languages.}
We show the size of the training data sets and name a few of the \numgrnlang{} languages.
} 
\label{fig:data-grnhrs}
\end{figure}
}

\newcommand{\insertMMSHours}{
\begin{figure}[t]
  \centering
\includegraphics[width=\linewidth]{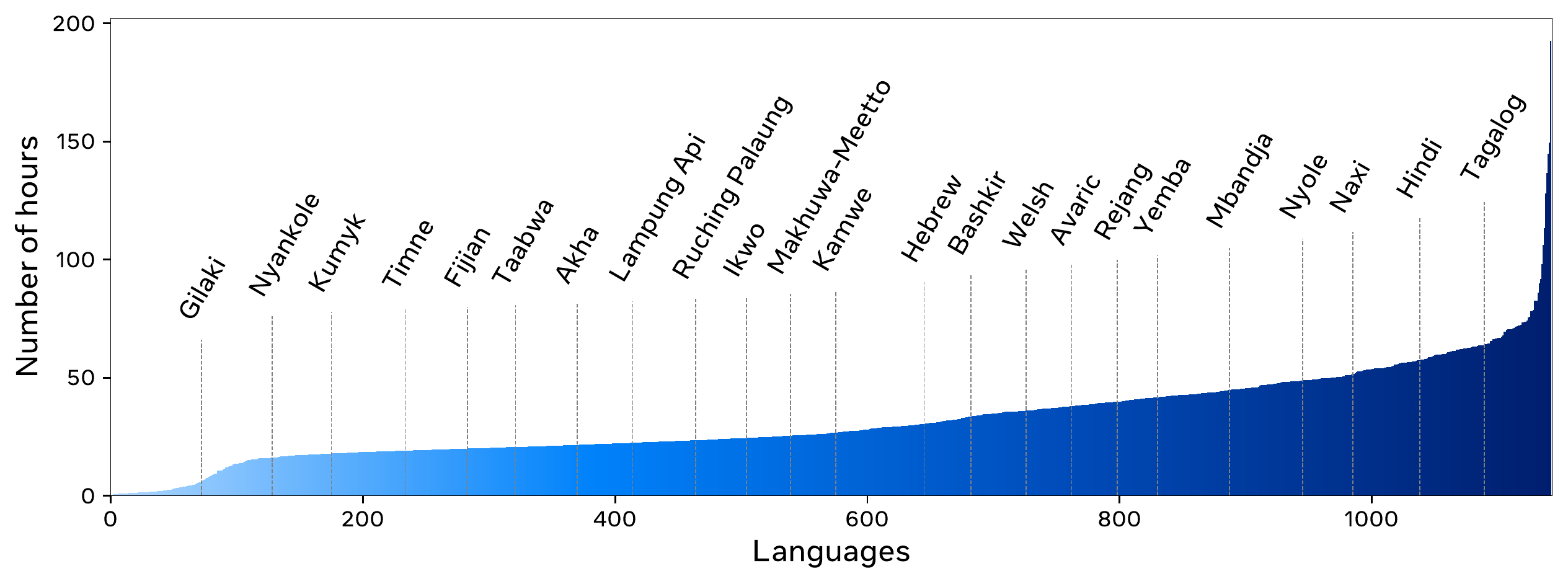}
\caption{
\textbf{\MMS{}: Amount of Speech Data across Languages.}
We show the size of the training data sets and name some of the \nummmslang{} languages.
}
\label{fig:data-mmshrs}
\end{figure}
}

\newcommand{\insertPretrainSetup}{
\begin{table*}[t]
\begin{center}
\resizebox{1\linewidth}{!}{
\begin{tabular}[b]{lrlrrrrrrr}
\toprule
Model & \#langs & Datasets & $B$ & $M$ & $F$ & $A$ & \#params \\
\midrule
\textit{Prior work} \\
XLSR-53 & 53 & MLS, CV, BBL & 24 & 1024 & 4096 & 16 & 317M \\
VP-100K & 23 & VP-100K & 24 & 1024 & 4096 & 16 & 317M \\
XLS-R (0.3B) & 128 & VP-400K, MLS, CV, VL, BBL & 24 & 1024 & 4096 & 16 & 317M \\
XLS-R (1B)   & 128 & VP-400K, MLS, CV, VL, BBL & 48 & 1024 & 4096 & 16 & 965M \\
\midrule
\modelnameb{0.3} & \numptlang{} & \MMS{}, FL, VP-400K, MLS, CV, VL, BBL & 24 & 1024 & 4096 & 16 & 317M \\
\modelnameb{1} & \numptlang{} & \MMS{}, FL, VP-400K, MLS, CV, VL, BBL & 48 & 1024 & 4096 & 16 & 965M \\
\bottomrule
\end{tabular}
}
\caption{\textbf{Self-supervised Models.}
Details of our models including prior work: XLSR-53~\citep{conneau2020unsupervised}, VP-100K~\citep{wang2021voxpopuli}, XLS-R~\citep{babu22_interspeech} and the \modelname{} models: 
the number of languages (\#langs), pretraining data (Datasets), the number of Transformer blocks ($B$), the number of hidden states ($M$), the inner dimension of feed-forward blocks ($F$), the number of attention heads ($A$) and the total number of parameters (\#params).
\label{tab:pretrain-models}}
\end{center}
\vspace{-0.4cm}
\end{table*}
}

\newcommand{\insertPretrainComparisonASR}{
\begin{figure}
\centering
\begin{tikzpicture}
\begin{axis}[
    ybar,
    bar width=.25in,
    width=.5*\textwidth,
    height=.35\textwidth,
    legend style={at={(0.95,0.95)},draw=none},
    nodes near coords align={vertical},
    ylabel={CER},
    ymin=12, 
    ymax=18,
    symbolic x coords={300M, 1B, 2B},
    enlarge x limits=0.4,
    xtick=data]
    \addplot[fill=metagray, pattern= north east lines] coordinates {
        (300M, 16.96)
        (1B, 13.92)
    };
    \addplot[fill=metablue] coordinates {
        (300M, 16.40)
        (1B, 13.20) 
    };
    
    \legend{XLS-R, \modelname{}}
\end{axis}
\end{tikzpicture}
\caption{\textbf{\modelname{} vs. XLS-R.} 
Character error rate (CER) on 61 FLEURS languages when fine-tuning multilingual ASR models on \MMS{} data. 
We report average performance on FLEURS dev data. 
}
\label{fig:pretrain-comparison-asr}
\end{figure}
}

\newcommand{\insertResultsASRLangScaling}{
\begin{center}
\begin{figure}[t]
\centering
\begin{tikzpicture}
\begin{axis}[
width=0.7\columnwidth,
height=.4\columnwidth,
legend style={font=\small,%fill=none,
at={(1.0,0.65)},
anchor=east,legend columns=1,draw=none,legend columns=2},
legend cell align={left},
xtick={61,128,256,512,1107},
x tick label style={xshift={(\ticknum==0)*-0.6em}},
x tick label style={xshift={(\ticknum==1)*0.2em}},
x tick label style={xshift={(\ticknum==2)*0.1em}},
ylabel={Character Error Rate},
ylabel near ticks, %ylabel shift={-10pt}
xlabel={Number of Languages supported by Model},
xlabel near ticks,
]
\addplot[blue,dashed,line width=1,,mark=+,mark options={scale=1}] coordinates {
    (61, 15.46)
    (128, 18.84)
    (256, 20.63)
    (512, 19.93)
    (1107, 20.69)
};
\addplot[metablue,line width=1,mark=+,mark options={scale=1}] coordinates {
    (61, 13.2)
    (128, 13.29)
    (256, 13.74)
    (512, 13.57)
    (1107, 13.63)
};
\addplot[black,dashdotted,line width=1,mark=+,mark options={scale=1}] coordinates {
    (61, 9.0)
    (128, 9.44)
    (256, 10.44)
    (512, 10.45)
    (1107, 11.07)
};
\addplot[metagray,loosely dashdotted,line width=1,mark=+,mark options={scale=1}] coordinates {
    (61, 7.47)
    (128, 7.34)
    (256, 7.72)
    (512, 7.7)
    (1107, 7.63)
};
\legend{FLEURS-61 (Dense), FLEURS-61 (LSAH), CV-49 (Dense), CV-49 (LSAH)}
\end{axis}
\end{tikzpicture}
\caption{
\textbf{Scaling Multilingual ASR to \nummmslang{} Languages.}
We fine-tune both dense and LSAH models with 61, 128, 256, 512 and \nummmslang{} languages on \MMS{} data and show average CER on 61 languages of FLEURS and 49 languages of CommonVoice. 
LSAH models have language-specific adapter modules and output layers.
}
\label{fig:results_asr_scale_lang}
\end{figure}
\end{center}
}

\newcommand{\insertResultsLIDFLVLCompare}{
\begin{figure}
\centering
\begin{tikzpicture}
\begin{axis}[
    ybar,
    bar width=.12in,
    width=0.9\linewidth,
    height=2in,
    legend style={fill=none, at={(0.96,0.92)},draw=none},
    nodes near coords align={vertical},
    scaled y ticks = false,
    legend columns=1,
    ylabel={Dev Accuracy},
    xlabel={Evaluation Set},
    ymin=70,
    xmax=3.5,
    xtick={1, 2},
    legend cell align={left},
    xticklabels={FLEURS, VoxLingua-107},
    enlarge x limits=0.2,
    xtick=data]
    \addlegendimage{empty legend}
   \addlegendentry[yshift=10pt]{Model Trained on}
    \addplot[fill=lightgray, pattern=] coordinates {
        (1, 98.46)
        (2, 91.51)
    };	
    \addlegendentry{FLEURS}
    \addplot[fill=metagray] coordinates {
        (1, 98.09)
        (2, 99.05)
    };
    \addlegendentry{VoxLingua-107}
    \addplot[fill=metablue, pattern=dots] coordinates {
        (1, 91.58)
        (2, 88.43)
    };
    \addlegendentry{\MMSU{}}
    \addplot[fill=metablue, pattern=north east lines] coordinates {
       (1,  92.01)
       (2, 84.23)
    };
    \addlegendentry{\GRN{}}
    \addplot[fill=metablue] coordinates {
       (1, 95.94)
       (2, 89.92)
    };
    \addlegendentry{\MMSUboth{}}
    \addplot[black,sharp plot,dashed] coordinates {(0.85,98.09) (1.33,98.09)};
    \addplot[black,<->,sharp plot,dashed] coordinates {(1.26,98.09)(1.26,95.94)};
    \node[font=\scriptsize] at (axis cs: 1.41,97) {-2.1\%};
    \addplot[black,sharp plot,dashed] coordinates {(1.7,91.51) (2.33,91.51)};
    \addplot[black,<->,sharp plot,dashed] coordinates {(2.26,91.51)(2.26,89.92)};
    \node[font=\scriptsize] at (axis cs: 2.43,90.8) {-1.6\%};    
\end{axis}
\end{tikzpicture}
\caption{
\textbf{LID Performance with Different Training Datasets.} 
Models trained on \MMSUboth{} are very competitive to models trained on existing data (FLEURS, VoxLingua-107) when evaluated out-of-domain (comparison in dashed line).
We train models on data from \MMSU{}, \GRN{}, FLEURS and VoxLingua-107 on a common subset of 72 languages and evaluate on FLEURS and VoxLingua-107.
}
\label{fig:lid-fl-vl-comp2}
\end{figure}
}

\newcommand{\insertLIDResultsScaling}{
\begin{table*}[t]
\begin{small}
\centering
\begin{tabular}{lrrrrrr}
\toprule
& \#lang & \multicolumn{2}{c}{in-domain} & \multicolumn{2}{c}{out-of-domain}\\
\cmidrule(lr){3-4}\cmidrule(lr){5-6}
 &  &  FLEURS & VL &  BABEL & VoxPopuli  \\ 
&  &   (102 lang.) &  (33 lang.) & (23 lang.) & (25 lang.) \\ 
\midrule
\multicolumn{2}{l}{\textit{Prior Work}} \\
mSLAM~\citep{Bapna2022mSLAMMM} & 102 & 77.7 & - & - & -  \\
Whisper~\citep{radford2022whisper}  & 82 & 64.5 & - & - & -  \\
ASRL~\citep{chen2023improving} & 102 & 95.9 & - & - & -  \\
XLS-R~\citep{babu22_interspeech}  & 107 & - & 94.3 & - & -  \\
SpeechBrain~\citep{speechbrain}  & 107 & - & 93.3 & - & -  \\
AmberNet~\citep{jia2022ambernet}  & 107 & - & 95.3 & - & -  \\
\midrule
\multicolumn{2}{l}{\textit{Our Baselines (based on existing datasets)}} \\
\modelname{} (FL) & 102 & 96.2 & - & - & - \\
\modelname{} (VL) & 107 & - & 94.7 & - & - \\
\modelname{} (FL + VL) & 126 & 97.4 & 94.3 & 78 &	87.8 \\
\midrule
\multicolumn{2}{l}{\textit{This Work}} \\
\modelname{} (\MMSUboth{}+FL+VL) & 126 & 97.5 &	93.9 & 84.1  & 87.3 \\
& 256 & 97.2 &	93.4 & 80.1 & 87.6 \\
& 512 & 96.8 &	92.9 & 81.6 & 85.6\\
& 1,024 & 97 & 92.8 & 80.5 & 86.2 \\
& 2,048 & 97.3 & 92.8 & 81.5 & 86.6 \\
&  \numlidlang{} & 97.2  & 93.9	 & 80.5&  87.1   \\
\bottomrule
\end{tabular}
\caption{
\textbf{Scaling LID to \numlidlang{} Languages.}
We show test accuracy of LID models trained on an increasing number of languages using data from FLEURS (FL), VoxLingua-107 (VL) and \MMSUboth{}; smaller language subsets are included in the larger subsets. 
There is little performance degradation when scaling to more languages.
We also show results from the literature as well as baselines trained only on FL or VL.} 
\label{tab:lid-scale}
\end{small}
\end{table*}
}

\newcommand{\insertDataRAIWords}{

\pgfplotstableread[row sep=\\,col sep=&]{
lang & mmstr & cctr & mmsout & flout \\
amh & 12.48 & 3.86 & 5.42 & 3.93\\
ara & 12.29 & 3.89 & 4.23 & 3.62\\
asm & 21.93 & 4.25 & 5.76 & 4.53\\
azj & 13.41 & 3.13 & 4.08 & 3.80\\
ben & 22.76 & 3.92 & 8.56 & 7.98\\
bul & 13.04 & 3.78 & 4.26 & 4.04\\
cat & 12.69 & 3.57 & 4.36 & 4.06\\
ceb & 19.23 & 5.94 & 6.69 & 6.32\\
cym & 16.32 & 4.51 & 5.60 & 4.81\\
deu & 22.40 & 3.50 & 8.37 & 8.18\\
ell & 15.32 & 3.59 & 5.60 & 3.41\\
eng & 20.31 & 5.17 & 6.93 & 6.66\\
fas & 21.52 & 6.14 & 9.25 & 6.78\\
fin & 16.25 & 3.57 & 4.65 & 4.60\\
fra & 15.27 & 4.75 & 3.96 & 3.44\\
guj & 18.38 & 4.33 & 4.79 & 4.11\\
hau & 25.44 & 7.78 & 10.63 & 7.81\\
heb & 13.67 & 3.64 & 4.25 & 3.72\\
hin & 15.35 & 3.62 & 4.23 & 3.43\\
hun & 19.45 & 3.63 & 6.63 & 6.65\\
ind & 23.56 & 6.15 & 6.35 & 6.28\\
jav & 29.96 & 5.61 & 9.23 & 8.88\\
kan & 12.66 & 2.90 & 4.84 & 4.29\\
kaz & 13.27 & 3.86 & 5.84 & 5.32\\
kir & 12.48 & 4.01 & 3.82 & 3.39\\
kor & 9.94 & 2.31 & 3.12 & 2.55\\
lav & 20.52 & 4.95 & 6.47 & 6.21\\
mar & 18.40 & 4.04 & 5.71 & 4.95\\
mon & 18.50 & 2.97 & 5.61 & 3.63\\
nld & 13.95 & 2.85 & 4.45 & 4.00\\
nya & 18.12 & 5.27 & 5.63 & 5.03\\
ory & 21.16 & 4.50 & 8.13 & 6.95\\
pan & 26.67 & 6.58 & 8.49 & 6.93\\
pol & 10.49 & 2.71 & 3.18 & 2.87\\
ron & 19.95 & 5.48 & 7.59 & 7.05\\
rus & 18.48 & 3.90 & 7.71 & 7.70\\
sna & 14.13 & 3.32 & 4.10 & 3.21\\
som & 16.89 & 3.75 & 6.44 & 5.46\\
spa & 10.07 & 2.08 & 2.38 & 2.13\\
swe & 16.22 & 3.97 & 5.26 & 4.43\\
swh & 13.87 & 3.81 & 3.75 & 3.70\\
tam & 9.21 & 2.08 & 3.64 & 3.33\\
tel & 8.79 & 2.20 & 4.18 & 3.15\\
tgk & 22.20 & 6.79 & 8.64 & 8.35\\
tgl & 10.76 & 3.00 & 2.67 & 2.57\\
tur & 8.79 & 2.18 & 2.98 & 2.25\\
ukr & 18.04 & 4.57 & 7.13 & 6.40\\
urd & 19.99 & 6.96 & 7.70 & 6.42\\
vie & 31.22 & 6.88 & 7.12 & 6.21\\
yor & 34.67 & 9.22 & 16.32 & 14.72\\
zlm & 24.54 & 6.06 & 7.26 & 6.68\\
}\dataRAIWords 
\begin{figure}
\centering
\begin{tikzpicture}
\begin{axis}[
width=1*\textwidth,
height=.4\textwidth,
ylabel={Cumulative Relative Frequency (\%)},
ylabel style={align=center,text width=5cm},
legend style={%font=\small,
at={(0.88,0.87)},
anchor=east,legend columns=4,draw=none},
xticklabels from table={\dataRAIWords}{lang},
x tick label style={rotate=45,anchor=east,font=\tiny},
enlarge x limits=0,
xtick=data]
\addplot[dashed,black,mark=square*,mark options={solid,scale=0.5,fill=black}] table[xticklabels from table={\dataRAIWords}{lang},x expr=\coordindex,x=lang,y=mmstr]{\dataRAIWords};
\addplot[solid,metablue,mark=diamond*,mark options={solid,scale=0.5,fill=metablue}] table[xticklabels from table={\dataRAIWords}{lang},x expr=\coordindex,x=lang,y=mmsout]{\dataRAIWords};
\addplot[dashdotdotted,red,mark=otimes,mark options={solid,scale=0.5,fill=red}] table[xticklabels from table={\dataRAIWords}{lang},x expr=\coordindex,x=lang,y=flout]{\dataRAIWords};
\legend{\MMS{} train, \MMS{} ASR output, FLEURS ASR output}
\end{axis}
\end{tikzpicture}
\caption{
\textbf{Analysis of Language Produced by \modelname{} ASR Models.}
\modelname{} models generate biased words at a slightly increased rate of 0.7\% absolute compared to models trained on FLEURS data.
We focus on words that occur at least twice as often in the \MMS{} training data compared to Common Crawl (in terms of relative frequency).
We compare the cumulative relative frequency of these words in the \modelname{} model output and the output of ASR models trained on FLEURS data.
ASR outputs are based on transcribing the development sets of 51 languages.
}
\label{fig:data-rai-words}
\end{figure}
}

\newcommand{\insertComparisonWithOtherDatasets}{
\begin{figure}
\centering
\begin{minipage}[t]{0.4\textwidth}
\begin{tikzpicture}
\begin{axis}[
    ybar,
    enlargelimits=0.1,
    bar width=.08in,
    width=\textwidth,
    height=2in,
    legend style={at={(0.83,0.95)},draw=none},
    legend cell align={left},
    nodes near coords align={vertical},
    ylabel={CER},
    ymin=5,
    symbolic x coords={eng, por, tel},
    legend style={fill=none, nodes={scale=0.9, transform shape}}, 
    x tick label style={rotate=45,font=\scriptsize}, 
    enlarge x limits=0.15,
    xtick=data]
    \addplot[fill=black, pattern=north east lines] coordinates {
        (eng, 17.16)
        (por, 10.49)
        (tel, 18.71)
    };
    \addplot[fill=metablue] coordinates {
        (eng, 12.44)
        (por, 8.256)
        (tel, 16.60)
    };    
    \legend{CMU Wilderness, {\MMS}}
\end{axis}
\end{tikzpicture}
\caption{
\textbf{{\MMS} vs. CMU Wilderness.} 
Character Error Rate of ASR models in English (eng), Portuguese (por) and Telugu (tel) on the FLEURS dev set.
}
\label{fig:data-compare-cmu}
\end{minipage}\hfill
\begin{minipage}[t]{0.55\textwidth}
\begin{tikzpicture}
\begin{axis}[
    width=1.05\linewidth,
    height=2in,
    legend style={fill=none, at={(0.67,0.97)}, draw=none},
    legend cell align={left},
    nodes near coords align={vertical},
    scaled y ticks = false,
    ylabel={CER},
    ymin=0,
    symbolic x coords={spa,cat,pol,deu,por,rus,ukr,tur,swh,fra,nld,fas,eng,cym,lug,ara,tam,tha},
    enlarge x limits=0.04,
    x tick label style={rotate=45,font=\scriptsize}, 
    xtick=data]
     \addplot[black,line width=0.6,mark=square*,mark options={scale=0.5}, dashed] coordinates {
(spa, 4.891304347826087)
(cat, 5.725425074907043)
(pol, 5.801607040367323)
(deu, 5.9194078623594955)
(por, 6.664313863890359)
(rus, 7.034013011395887)
(ukr, 7.865942118075193)
(tur, 7.907471931862176)
(swh, 7.922350472193075)
(fra, 8.30560720227028)
(nld, 9.047417442845047)
(fas, 9.172106615134448)
(eng, 9.468897855518914)
(cym, 12.19765442816048)
(lug, 12.924718556148367)
(ara, 15.041513805754006)
(tam, 15.646954117669337)
(tha, 16.505637617410347)
    };				
    \addplot[metablue,line width=0.6,mark=*,mark options={solid, scale=0.5}] coordinates {
(spa, 6.516737796745799)
(cat, 7.880581928450236)
(pol, 7.891716089535106)
(deu, 9.09466721785862)
(por, 8.164225633786248)
(rus, 9.717067633061172)
(ukr, 11.485406575032581)
(tur, 9.388308168795973)
(swh, 11.60620596612202)
(fra, 11.689010666405714)
(nld, 12.320067739204065)
(fas, 13.180779110597884)
(eng, 13.477254572639945)
(cym, 17.19859061430833)
(lug, 15.717458453526664)
(ara, 14.771191349681406)
(tam, 17.91171959325115)
(tha, 22.498607402854343)

    };
    \legend{Common Voice, \MMS{}}
\end{axis}
\end{tikzpicture}
\caption{\textbf{{\MMS} vs. Common Voice.}
Character Error Rate on FLEURS dev set for ASR models trained on Common Voice (CV) data and {\MMS} data for 18 languages. }
\label{fig:compare-cv}
\end{minipage}

\end{figure}
}

\newcommand{\insertUromanXsampaExamples}{
\begin{table}
\centering
\includegraphics[width=0.75\linewidth]{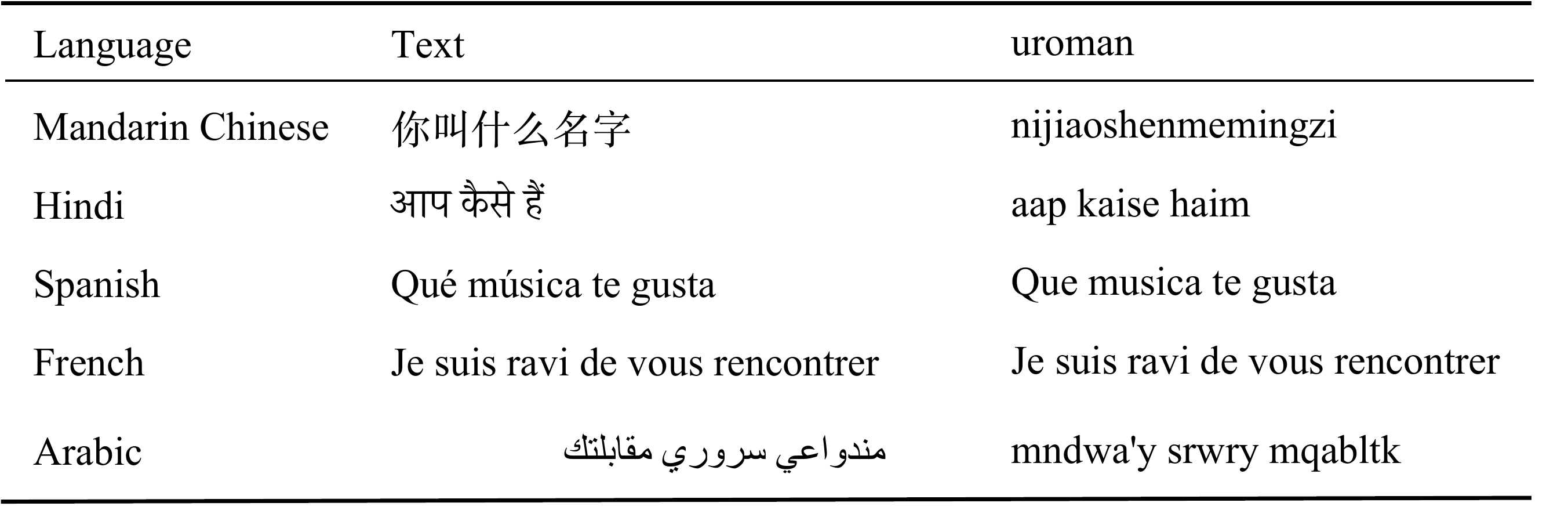}
\caption{
\textbf{Illustration of Text Encoding for Forced Alignment.}
Example outputs of uroman~\citep{hermjakob-etal-2018-box}.
} 
\label{tab:data-ex-uroman-xsampa}
\end{table}

}

\newcommand{\insertForceAlignBenchmark}{
\begin{figure}[t]
\centering
\begin{tikzpicture}
\begin{axis}[
width=0.6\textwidth,
height=0.4\textwidth,
xtick={0, 100, 500, 1000, 2000},
x tick label style={xshift={(\ticknum==1)*0.7em}},
xmin=0,
ymin=-2,
ymax=100,
ylabel={Wall clock time (sec)},
ylabel near ticks, %ylabel shift={-10pt}
xlabel={Input audio length (sec)},
xlabel near ticks,
log ticks with fixed point,
grid style={dotted,lightgray}
grid=both,
legend style={fill=none, draw=none, at={(0.5,0.91)},anchor=north},
legend cell align={left},
]

\addplot[metagray,line width=1,mark=square*,mark options={scale=0.75}, dashed] coordinates {
    (5, 0.060311791)
    (100, 3.9413602381)
    (500, 94.003602301)
    (1000, 461.9733619)
    (2000, 3080.95882)
};
\addplot[metaorange,line width=1.2,mark=triangle*,mark options={scale=1},dashdotted] coordinates {
    (5, 0.0002211)
    (100, 0.228787)
    (500, 5.66702)
    (1000, 22.162)
    (2000, 89.3203)
};

\addplot[metablue,line width=1,mark=*,mark options={scale=0.75}] coordinates {
    (5, 0.00272785)
    (100, 0.075407437682)
    (500, 1.0982227040920)
    (1000, 3.683633243166)
    (2000, 14.5421336991)
};
\legend{ctc-seg (CPU), Flashlight (CPU), MMS (GPU) }
\end{axis}
\end{tikzpicture}
\captionof{figure}{
\textbf{Efficiency of Forced Alignment Implementations.}
The \modelname{} implementation runs on GPU and can process long audio sequences in reasonable time compared to CPU alternatives.
}
\label{fig:results_fa_benchamrk}
\end{figure}

}

\newcommand{\insertStarPlots}{
\begin{figure}
\centering
\begin{subfigure}[l]{.47\textwidth}
\includegraphics[width=0.99\textwidth,trim={1cm 1cm 1cm 1cm}]{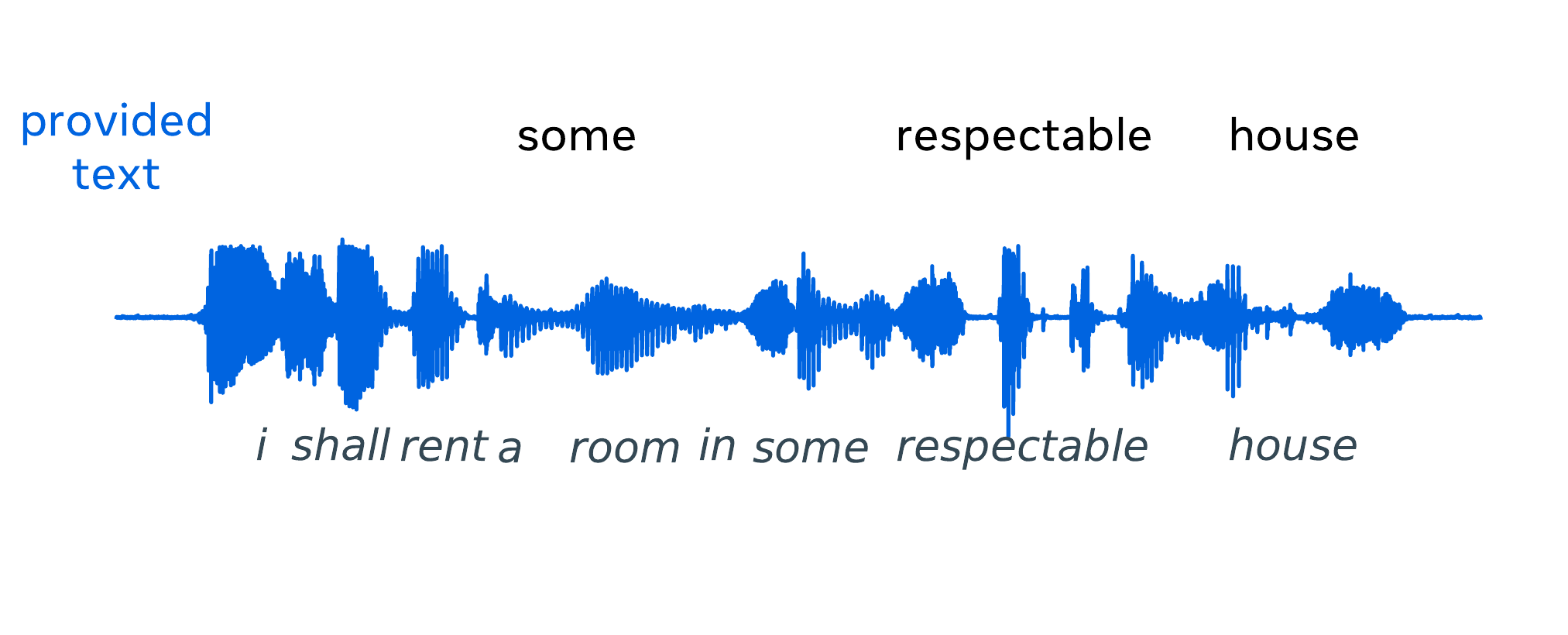}
\end{subfigure}
\begin{subfigure}[l]{.47\textwidth}
\includegraphics[width=0.99\textwidth,trim={1cm 1cm 1cm 1cm}]{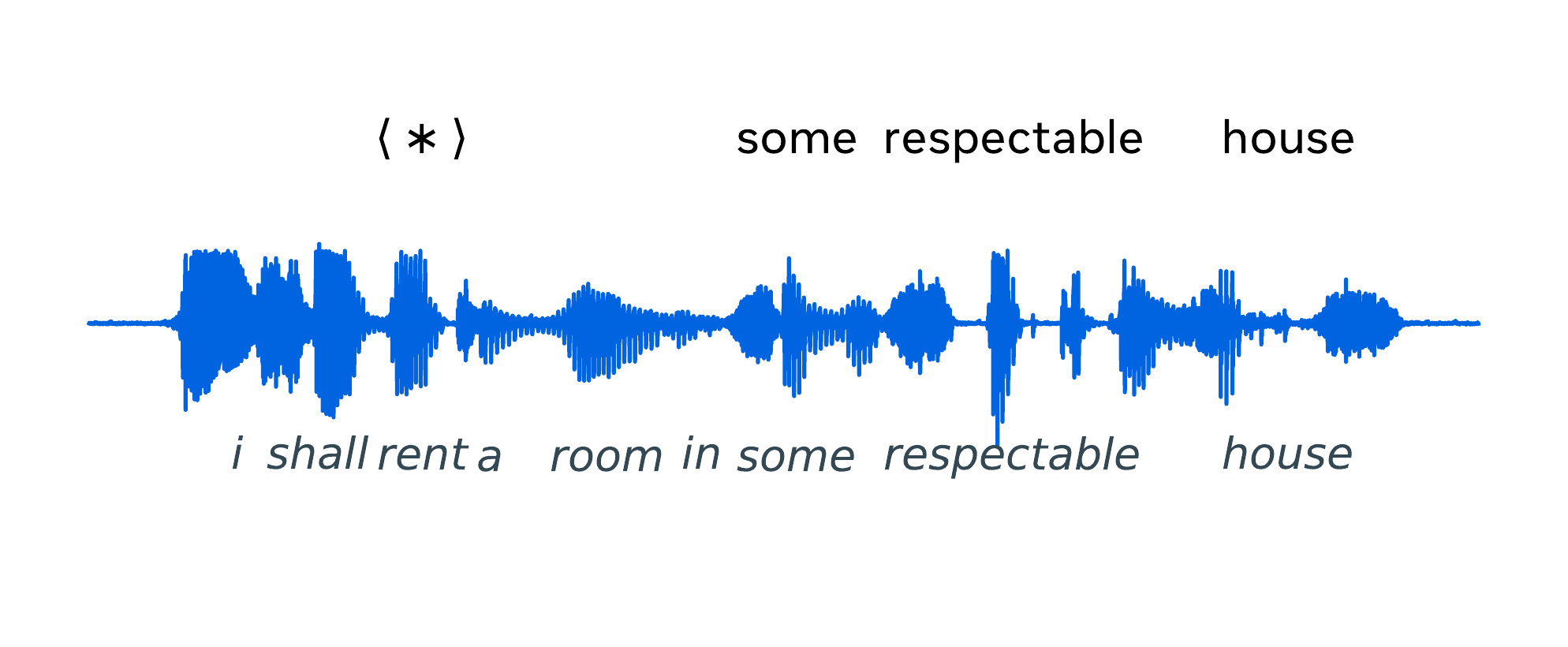} 
\end{subfigure}
\caption{
\textbf{Illustration of the $\Star$ Token in Forced Alignment.}
We show the text to which we would like to align the audio at the top and what is actually spoken at the bottom in italics.
Left: erroneous alignment when the provided text is incomplete. The word \emph{some} is aligned incorrectly.
Right: using $\Star$ token at the beginning enables correct alignment of the provided text.
}
\label{fig:star_token}
\end{figure}
}

\newcommand{\insertTableResultsTTSDesign}{
\begin{table}[t]
\centering
\begin{tabular}{lll|rrr|rrr}
\toprule
train & train & text & \multicolumn{3}{c}{ASR (CER)} &  \multicolumn{3}{|c}{MOS} \\
upd & data & repr. & \MMS{} & LJS & FLEURS & \MMS{} & LJS & FLEURS \\
\midrule
\multicolumn{3}{l|}{\emph{Natural speech}} & 4.4 & 4.3 & 9.3 & 3.89	\ci{0.06} & 3.96 \ci{0.06} & 3.37 \ci{0.07} \\
\midrule
800K & LJS & phon. & 5.5 & 4.9 & 5.9 & 3.87 \ci{0.08} & 3.82 \ci{0.06} & 3.73 \ci{0.07} \\
100K & LJS & phon. & 6.3 & 4.9 & 6.3 & 3.64 \ci{0.09} & 3.74 \ci{0.07} & 3.66 \ci{0.07} \\
100K & \MMS{} & phon. & 7.2 & 6.8 & 7.9 & 3.68 \ci{0.07} & 3.51 \ci{0.08} & 3.54 \ci{0.07} \\
100K & \MMS{} & chars & 7.2	& 9.2 & 10.0 & 3.58 \ci{0.08} & 3.45 \ci{0.08} & 3.34 \ci{0.09} \\
\bottomrule
\end{tabular}
\vspace{1mm}
\caption{
\textbf{Ablation of Training Setup.} 
Our design choices (last row) lead to a moderate quality reduction compared to the standard VITS setup (second row) but enable scaling TTS to \nummmslang{} languages.
The standard VITS setting performs 800K training updates on clean LJSpeech data using a high quality phonemizer while as \modelname{} is trained on \MMS{} data for fewer updates and with characters.
Natural speech results are based on the human utterances of each dataset.
We report ASR CER and MOS scores from a human study with confidence interval 95\% on the English development sets of \MMS{}, LJSpeech and FLEURS.
} 
\label{tab:tts-design}
\end{table}
}

\newcommand{\insertTableResultsTTSDrama}{
\begin{table}[t]
\centering
\begin{tabular}{l|rrr|rrr}
\toprule
& \multicolumn{3}{c|}{ASR (CER)} & \multicolumn{3}{c}{MOS} \\
& \MMS{} & LJS & FLEURS & \MMS{} & LJS & FLEURS \\
\midrule
\emph{Natural speech} & 4.4 & 4.3 & 9.3 & 3.89 \ci{0.06} & 3.96 \ci{0.06} & 3.37 \ci{0.07} \\
no background music & 7.2 & 9.2 & 10.0 & 3.51 \ci{0.07} & 3.52 \ci{0.08} & 3.41 \ci{0.09} \\
\midrule
background music & 11.8 & 15.6 & 16.1 & 3.24 \ci{0.08} & 2.98 \ci{0.08} & 3.01 \ci{0.08} \\
\hspace{2mm}+ denoise & 10.8 & 13.2 & 14.4 & 3.32 \ci{0.08} & 3.18 \ci{0.07} & 3.16 \ci{0.08} \\
\hspace{2mm}+ denoise + filter & 7.8 & 10.8 & 11.9 & 3.47 \ci{0.07} & 3.32 \ci{0.08} & 3.12 \ci{0.07} \\
\bottomrule
\end{tabular}
\vspace{1mm}
\caption{
\textbf{Ablation of Building TTS Models using Data with Background Music.}
Curating the training data containing background music results in TTS systems which approach the performance of models trained on recordings without background music. 
We show MOS ratings as well as ASR CER on three different benchmarks for English data for the development sets of \MMS{}, LJSpeech and FLEURS.
The TTS model labeled "no background music" is identical to the last row in~\autoref{tab:tts-design} and we use the \MMS{} development set of the non-drama recording.
} 
\label{tab:tts-drama}
\end{table}
}

\newcommand{\insertTableResultsTTSFLEURS}{
\begin{table}
\centering
\begin{tabular}{lrrrr}
\toprule
& \multicolumn{2}{c}{ASR (CER)} & \multicolumn{2}{c}{MOS} \\
& TTS & ref & TTS & ref \\
\midrule
In-domain     & 11.1 & 9.2 & 3.51 {\tiny $\pm$ 0.11} & 3.61 {\tiny $\pm$ 0.11} \\
Out-of-domain & 11.3 & 8.8 & 3.52 {\tiny $\pm$ 0.11} & 3.33	{\tiny $\pm$ 0.12}$^*$ \\
\bottomrule
\end{tabular}
\vspace{1mm}
\caption{
\textbf{TTS in-domain and Out-of-domain Evaluation on 61 Languages.}
We synthesize the test sets of \MMS{} (in-domain) and FLEURS (out-of-domain) to measure character error rate (CER) and collect human judgements (MOS) for both the model outputs (TTS) and human utterances (ref). 
The \modelname{} models are robust with little performance degradation when applied to general domain data (out-of-domain):
they retain much of the original content (low difference between CER of TTS and ref) and the systems produce outputs with good prosody and sound quality (low difference between MOS scores between TTS and ref).
We show MOS scores with confidence interval 95\%.
($^*$) human judges assigned low ratings since the human reference audio in FLEURS contains a lot of variability compared to synthesized speech or the \MMS{} reference speech (in-domain ref).
} 
\label{tab:tts-fleurs}
\end{table}
}

\newcommand{\insertTableResultsTTSFull}{
\begin{table}
\centering
\begin{tabular}{lrrrrrr}
\toprule
& \#lang & MCD & \multicolumn{2}{c}{ASR (CER)} & TTS CER & \% \\
& & & TTS & ref & $\leq$ 5 & \\
\midrule
Asia & 335 & 4.30 \ci{0.1} & 3.1 \ci{0.2} & 1.9 \ci{0.1} & 296 & 88\% \\
South America & 136 & 4.10 \ci{0.1} & 2.6 \ci{0.2} & 1.8 \ci{0.1} & 129 & 95\% \\
North America & 144 & 4.12 \ci{0.1} & 3.8 \ci{0.8} & 2.4 \ci{0.2} & 125 & 87\% \\
Europe & 41 & 4.33 \ci{0.2} & 3.0 \ci{0.3} & 1.9 \ci{0.2} & 39 & 95\% \\
Africa & 363 & 4.34 \ci{0.1} & 4.1 \ci{0.2} & 2.6 \ci{0.1} & 277 & 76\% \\
Pacific & 88 & 4.72 \ci{0.2} & 3.4 \ci{1.3} & 1.8 \ci{0.2} & 79 & 90\% \\
\midrule
& 1,107 & 4.30 \ci{0.0} & 3.5 \ci{0.2} & 2.2 \ci{0.1} & 945 & 85\% \\
\bottomrule
\end{tabular}
\vspace{1mm}
\caption{
\textbf{TTS Evaluation on \nummmslang{} Languages.}
The majority of \modelname{} TTS models can synthesize speech which preserves most of the content as per the ASR character error rates (TTS CER $\leq$ 5).
We report MCD and character error rate for the synthesized outputs (TTS) of the \MMS{} test sets as well as the human references (ref).
We also show the number of systems which achieve CER less than five, indicating systems which on average produce no more than one incorrect character every twenty characters.
Results are shown with confidence interval 95\%.
} 
\label{tab:tts-full}
\end{table}
}

\newcommand{\insertResultsASRWisper}{
\begin{table}[]
\centering
\begin{tabular}{lrrrr}
\toprule
& \#lang & labeled train & \multicolumn{2}{c}{FLEURS-54} \\
& & data (h) & dev & test \\
\midrule
\textit{Prior Work} \\
Whisper medium & 99 & 680K & - & 50.1 \\
Whisper large-v2 & 99 & 680K & - & 44.3 \\
\midrule
\textit{This Work} \\
\modelname{} & 61 & 3K & 20.9 & 20.7 \\
\modelname{} (LSAH) & 61 & 3K & 19.0 & 19.1 \\
\modelname{} & \nummmslang{} & \nummmshrsrd{} & 24.8 & 24.8 \\
\modelname{} (LSAH) & 1,107 & \nummmshrsrd{} & 18.7 & 18.7 \\
\bottomrule
\end{tabular}
\vspace{1mm}
\caption{
\textbf{Comparison to Whisper.} 
We report average WER on the 54 languages of the FLEURS benchmark supported by both Whisper and \modelname{} (FLEURS-54).
\modelname{} is a CTC-based model and to enable a fairer comparison we use n-gram models trained on web data when comparing to Whisper whose decoder is a neural sequence-model that serves as a language model and was trained on billions of web tokens. 
} 
\label{tab:asr-whisper} 
\end{table}
}

\newcommand{\insertResultsASRWisperFull}{
\begin{table}[t]
\centering
\begin{tabular}{lrrrrr}
\toprule
& \#lang & lbld train & \multicolumn{2}{c}{FLEURS-54} \\
& & data (h) & (M) & dev & test \\
\midrule
\textit{Prior Work} \\
Whisper medium & 99 & \multirow{2}{*}{680K} & 769M & - & 50.1 \\
Whisper large-v2 & 99 & & 1,550M & - & 44.3 \\
\midrule
\textit{This Work} \\
\modelname{} & 61 & 3K & 965M & 33.6 & 33.3 \\
\hspace{2mm} + CC LM & 61 & 3K & 965M & 20.9	& 20.7\\ 
\modelname{} (LSAH) & 61 & 3K & 1,096M & 31.4	& 31.0 \\
\hspace{2mm} + CC LM & 61 & 3K &  1,096M & 19.1 & 19.0\\
\midrule
\modelname{} & \nummmslang{} & \nummmshrsrd{} & 965M & 44.7 &	44.2 \\
\hspace{2mm} + CC LM  & \nummmslang{} & \nummmshrsrd{} & 965M & 24.8 &	24.8 \\
\modelname{} (LSAH) & 1,107 & \nummmshrsrd{} & 3,346M & 32.8  &	32.5 \\
\hspace{2mm} + CC LM  & 1,107 & \nummmshrsrd{} & 3,346M & 18.7 & 18.7 \\
\bottomrule
\end{tabular}
\vspace{1mm}
\caption{
\textbf{Comparison to Whisper.} 
We report average WER on the 54 languages of the FLEURS benchmark supported by both Whisper and \modelname{} (FLEURS-54).
\modelname{} is a CTC-based model and to enable a fairer comparison we use n-gram models trained on web data when comparing to Whisper whose decoder is a neural sequence-model and serves as a language model that was trained on billions of web tokens. 
} 
\label{tab:asr-whisper-full} 
\end{table}
}

\newcommand{\insertResultsASRWisperPerLang}{
\begin{table}[h!]
\centering
\begin{scriptsize}
\begin{tabular}{lrrrrrrrrrrrr}
\toprule
& Whisper & Whisper & \modelname{} & \modelname{} & \modelname{} & \modelname{} & \modelname{} & \modelname{} & \modelname{} & \modelname{} \\
& medium & large-v2 & L-61 & L-61 & L-61 & L-61 & L-1107 & L-1107 & L-1107 & L-1107 \\
&  &  & noLM & CC LM & noLM & CC LM & noLM & CC LM & noLM & CC LM \\
&  &  &  &  & LSAH & LSAH &  &  & LSAH & LSAH \\
\midrule
Amharic & 229.3 & 140.3 & 48.7 & 30.7 & 52.4 & 32.5 & 52.9 & 30.1 & 53.3 & 31.1\\
Arabic & 20.4 & 16.0 & 34.9 & 19.6 & 35.8 & 19.9 & 44.0 & 23.4 & 41.3 & 21.0\\
Assamese & 102.3 & 106.2 & 29.5 & 18.8 & 28.4 & 18.6 & 37.6 & 21.2 & 30.5 & 19.2\\
Azerbaijani & 33.1 & 23.4 & 40.7 & 21.3 & 38.3 & 19.8 & 45.0 & 21.2 & 40.1 & 19.1\\
Bengali & 100.6 & 104.1 & 19.7 & 11.6 & 20.0 & 12.1 & 25.0 & 12.5 & 23.5 & 12.1\\
Bulgarian & 21.4 & 14.6 & 23.4 & 13.1 & 23.9 & 13.3 & 27.9 & 12.9 & 25.5 & 13.5\\
Burmese & 123.0 & 115.7 & 22.2 & 14.2 & 22.3 & 14.5 & 29.2 & 20.2 & 24.5 & 16.0\\
Catalan & 9.6 & 7.3 & 18.1 & 11.0 & 18.1 & 11.0 & 25.9 & 11.5 & 20.1 & 10.8\\
Dutch & 9.9 & 6.7 & 26.9 & 13.7 & 26.4 & 14.3 & 38.1 & 14.9 & 27.6 & 14.5\\
English & 4.4 & 4.2 & 23.8 & 10.7 & 24.8 & 11.8 & 38.8 & 12.2 & 27.8 & 12.3\\
Filipino & 19.1 & 13.8 & 19.3 & 11.9 & 19.4 & 12.2 & 26.2 & 13.5 & 20.2 & 12.4\\
Finnish & 13.9 & 9.7 & 26.4 & 22.5 & 26.9 & 23.1 & 32.3 & 22.2 & 28.8 & 23.1\\
French & 8.7 & 8.3 & 24.3 & 13.7 & 24.5 & 14.1 & 35.8 & 15.4 & 29.3 & 15.0\\
German & 6.5 & 4.5 & 22.5 & 13.2 & 22.3 & 13.7 & 38.4 & 13.1 & 22.5 & 13.3\\
Greek & 19.0 & 12.5 & 40.8 & 14.0 & 40.5 & 13.6 & 57.5 & 13.0 & 40.1 & 13.6\\
Gujarati & 104.8 & 102.7 & 23.0 & 13.0 & 22.7 & 12.8 & 73.9 & 56.4 & 24.0 & 12.8\\
Hausa & 106.6 & 88.9 & 35.9 & 26.7 & 36.3 & 27.3 & 40.4 & 26.7 & 38.3 & 26.4\\
Hebrew & 33.1 & 27.1 & 68.5 & 44.8 & 66.6 & 41.5 & 78.7 & 50.9 & 67.1 & 40.0\\
Hindi & 26.8 & 21.5 & 65.0 & 44.4 & 28.8 & 16.0 & 70.7 & 45.7 & 21.2 & 10.6\\
Hungarian & 24.3 & 17.0 & 31.2 & 18.1 & 30.7 & 18.4 & 40.3 & 18.3 & 30.7 & 18.0\\
Icelandic & 49.9 & 38.2 & 42.9 & 18.3 & 42.3 & 19.9 & 53.6 & 20.5 & 45.3 & 18.6\\
Indonesian & 10.2 & 7.1 & 25.5 & 11.7 & 23.8 & 12.1 & 31.9 & 11.6 & 23.4 & 11.8\\
Javanese & 67.9 & 68.5 & 32.8 & 19.6 & 32.8 & 20.0 & 58.8 & 27.2 & 34.2 & 19.5\\
Kannada & 77.7 & 37.0 & 18.8 & 14.4 & 15.8 & 12.9 & 41.3 & 25.2 & 17.7 & 13.3\\
Kazakh & 48.8 & 37.7 & 30.2 & 17.4 & 30.2 & 17.7 & 63.8 & 19.5 & 31.6 & 17.4\\
Khmer & 103.8 & 128.9 & 26.0 & 19.9 & 25.7 & 19.8 & 70.7 & 52.4 & 26.7 & 19.7\\
Korean & 16.4 & 14.3 & 58.7 & 37.5 & 59.9 & 37.3 & 82.1 & 58.2 & 68.3 & 40.1\\
Lao & 101.4 & 101.5 & 48.9 & 45.4 & 24.2 & 22.8 & 62.1 & 56.6 & 22.6 & 16.9\\
Latvian & 32.0 & 23.1 & 20.8 & 12.0 & 20.9 & 12.1 & 24.5 & 11.9 & 21.8 & 12.1\\
Malay & 12.2 & 8.7 & 25.3 & 12.3 & 25.9 & 13.2 & 32.4 & 12.1 & 26.1 & 12.5\\
Malayalam & 101.1 & 100.7 & 23.7 & 19.1 & 19.5 & 16.6 & 39.1 & 25.6 & 20.4 & 15.3\\
Marathi & 63.2 & 38.3 & 32.5 & 19.0 & 19.2 & 13.5 & 28.0 & 14.9 & 20.9 & 13.4\\
Mongolian & 103.7 & 110.5 & 55.7 & 29.3 & 54.9 & 32.9 & 67.7 & 28.7 & 55.3 & 32.3\\
Persian & 41.0 & 32.9 & 39.7 & 22.9 & 39.9 & 22.5 & 44.4 & 21.3 & 42.9 & 22.0\\
Polish & 8.0 & 5.4 & 21.5 & 11.4 & 20.8 & 11.6 & 33.0 & 11.0 & 25.1 & 11.3\\
Portuguese & 5.0 & 4.3 & 16.1 & 10.8 & 16.3 & 10.8 & 19.3 & 10.2 & 17.7 & 10.5\\
Punjabi & 102.0 & 102.4 & 41.4 & 29.9 & 30.4 & 20.7 & 99.0 & 91.0 & 31.0 & 19.8\\
Romanian & 20.0 & 14.4 & 27.9 & 18.8 & 28.4 & 19.1 & 31.3 & 17.8 & 27.4 & 18.3\\
Russian & 7.2 & 5.6 & 30.3 & 14.6 & 30.4 & 14.3 & 38.8 & 14.7 & 35.0 & 15.0\\
Shona & 143.9 & 121.0 & 38.1 & 30.4 & 37.7 & 30.1 & 43.0 & 29.9 & 37.8 & 29.6\\
Somali & 104.0 & 102.9 & 51.8 & 42.8 & 52.5 & 43.0 & 54.5 & 42.9 & 53.8 & 42.8\\
Spanish & 3.6 & 3.0 & 12.2 & 7.8 & 12.4 & 8.2 & 14.0 & 7.8 & 14.0 & 8.7\\
Swahili & 52.8 & 39.3 & 22.9 & 16.0 & 23.3 & 15.6 & 29.6 & 16.8 & 23.7 & 16.0\\
Swedish & 11.2 & 8.5 & 29.9 & 17.4 & 30.5 & 17.5 & 38.2 & 17.2 & 33.5 & 17.4\\
Tajik & 74.0 & 85.8 & 59.8 & 46.6 & 33.9 & 19.2 & 59.0 & 39.5 & 25.7 & 15.7\\
Tamil & 23.1 & 17.5 & 24.2 & 18.3 & 21.9 & 16.3 & 25.3 & 17.3 & 23.9 & 16.3\\
Telugu & 82.8 & 99.0 & 19.4 & 13.7 & 19.6 & 13.7 & 24.5 & 15.8 & 22.1 & 13.6\\
Thai & 15.4 & 11.5 & 18.2 & 13.6 & 18.1 & 13.6 & 27.6 & 18.8 & 20.7 & 14.3\\
Turkish & 10.4 & 8.4 & 28.6 & 17.3 & 28.7 & 17.5 & 31.2 & 16.1 & 30.9 & 16.9\\
Ukrainian & 11.6 & 8.6 & 31.1 & 13.6 & 31.7 & 13.5 & 39.2 & 13.3 & 33.3 & 13.6\\
Urdu & 28.2 & 22.6 & 42.3 & 22.7 & 33.1 & 20.1 & 46.4 & 25.1 & 36.9 & 20.5\\
Vietnamese & 12.7 & 10.3 & 44.5 & 18.6 & 47.5 & 20.7 & 56.6 & 21.0 & 52.9 & 19.8\\
Welsh & 40.8 & 33.0 & 48.9 & 20.8 & 49.0 & 20.8 & 54.9 & 21.4 & 51.4 & 20.9\\
Yoruba & 105.1 & 94.8 & 61.2 & 49.7 & 61.9 & 49.4 & 62.7 & 50.2 & 64.2 & 49.4\\
\midrule 
& 50.1 & 44.3 & 33.3 & 20.7 & 31.0 & 19.1 & 44.2 & 24.8 & 32.5 & 18.7\\
\bottomrule
\end{tabular}
\end{scriptsize}
\vspace{1mm}
\caption{
\textbf{Comparison to Whisper on the FLEURS test set.}
We report WER for each of the 54 languages supported by both \modelname{} and Whisper, except for Thai (tha), Lao (lao), Burmese (mya) and Khmer (khm) where we report CER. 
We apply Whisper normalization for both reference and hypothesis for measuring CER/WER.
} 
\label{tab:asr-whisper-perlang-test} 
\end{table}
}

\newcommand{\insertResultsASRUSM}{
\begin{table}[t]
\centering
\begin{tabular}{lrr}
\toprule
& \multicolumn{2}{c}{FLEURS-102} \\
& dev & test \\
\midrule
\textit{Prior Work} \\
w2v-BERT~\citep{chen2022maestro} & - & 12.3 \\
Maestro-U~\citep{chen2022maestro} & - & 8.7 \\
USM~\citep{zhang2023usm} & - & 6.9 \\
USM-M~\citep{zhang2023usm} & - & 6.5 \\
USM-M-adapter~\citep{zhang2023usm} & - & 6.7 \\
\midrule
\textit{This Work} \\
\modelname{} FL-102 (LSFT) + LM & 6.3 & 6.3 \\
\bottomrule
\end{tabular}
\vspace{1mm}
\caption{
\textbf{Comparison to Google USM.} 
We report average CER on the 102 languages of FLEURS and use n-gram models together with our CTC acoustic models.
USM-M is an RNN-T model which uses both unlabeled text as well as labeled speech data during pre-training and for \modelname{} we use unlabeled text to train language models for inference.
}
\label{tab:asr-usm}
\end{table}
}

\newcommand{\insertPretainXLSRvsMMS}{

\pgfplotstableread[row sep=\\,col sep=&]{
lang & delta \\
amh & 7.94 \\
lao & 6.57 \\
mal & 5.63 \\
nya & 4.95 \\
swh & 2.62 \\
ful & 2.12 \\
kor & 2.11 \\
pan & 2 \\
lug & 1.97 \\
urd & 1.42 \\
khm & 1.41 \\
mar & 1.4 \\
hin & 1.38 \\
orm & 1.34 \\
tel & 1.27 \\
hau & 1.23 \\
tam & 1.23 \\
guj & 1.04 \\
vie & 0.87 \\
mya & 0.82 \\
jav & 0.76 \\
asm & 0.72 \\
azj & 0.69 \\
tha & 0.57 \\
som & 0.5 \\
ben & 0.43 \\
kaz & 0.39 \\
tgl & 0.28 \\
yor & 0.28 \\
heb & 0.22 \\
zlm & 0.21 \\
ceb & 0.2 \\
ind & 0.2 \\
tur & 0.17 \\
mon & 0.11 \\
ory & 0.11 \\
kan & 0.06 \\
fin & 0.02 \\
nld & -0.01 \\
sna & -0.03 \\
bul & -0.06 \\
ron & -0.14 \\
ara & -0.22 \\
ukr & -0.24 \\
pol & -0.24 \\
lav & -0.25 \\
kir & -0.26 \\
deu & -0.26 \\
cat & -0.28 \\
hun & -0.29 \\
swe & -0.36 \\
por & -0.39 \\
fas & -0.48 \\
fra & -0.51 \\
rus & -0.54 \\
eng & -0.64 \\
ell & -0.66 \\
spa & -0.7 \\
isl & -1.14 \\
tgk & -1.72 \\
cym & -1.87 \\
}\dataDelta 
\begin{figure}
\centering
\begin{tikzpicture}
\begin{axis}[
width=1*\textwidth,
height=.4\textwidth,
ylabel={CER \modelname{}} - XLS-R,
ylabel style={align=center,text width=5cm},
legend style={%font=\small,
at={(0.88,0.87)},
minor tick num=1,
anchor=east,legend columns=4,draw=none},
ymin=-5,
ymax=10,
xticklabels from table={\dataDelta}{lang},
x tick label style={rotate=45,font=\tiny},
enlarge x limits=0,
xtick=data]
\addplot[solid,metablue,line width=0.6,mark=*,mark options={solid, scale=0.4}] table[xticklabels from table={\dataDelta}{lang},x expr=\coordindex,x=lang,y=delta]{\dataDelta};
\addplot[color=gray,dashed] coordinates {(0,0) (61,0)};
\end{axis}
\end{tikzpicture}
\caption{
\textbf{\modelname{} vs. XLS-R Breakdown.}
We show the absolute character error rate difference between multilingual ASR models based on XLS-R and \modelname{} models with 1B parameters.
Positive values indicate better performance of \modelname{} and negative values better performance of XLS-R.
Models are fine-tuned on \MMS{} data and evaluated on the development sets of 61 FLEURS languages.
}
\label{fig:pretrain-xlsr-mms-perlang}
\end{figure}
}

\newcommand{\insertResultsGenderBiasFull}{
\begin{table}[h!]
\centering
\begin{tabular}{l|rr|rrr|rrr}
\toprule
& \multicolumn{2}{c}{Num Samples} & \multicolumn{3}{c}{\modelname{} ASR CER $\downarrow$} & \multicolumn{3}{c}{FLEURS ASR CER $\downarrow$}  \\
& Female & Male & Female & Male & Overall & Female & Male & Overall \\
\midrule
Assamese & 139 & 279 & 15.3 & 15.1 & 15.4 & 8.6 & 8.0 & 8.9 \\
Bulgarian & 189 & 206 & 9.0 & 10.2 & 7.9 & 3.6 & 3.5 & 3.6 \\
Welsh & 150 & 296 & 15.8 & 19.4 & 14.1 & 8.4 & 11.5 & 6.9 \\
Greek & 125 & 146 & 11.4 & 9.3 & 13.1 & 5.4 & 4.0 & 6.6 \\
English & 245 & 149 & 10.8 & 10.7 & 10.9 & 6.0 & 5.5 & 6.8 \\
Spanish & 130 & 278 & 5.2 & 5.8 & 5.0 & 2.3 & 2.9 & 2.0 \\
Fula & 107 & 166 & 27.2 & 29.0 & 26.2 & 14.8 & 21.4 & 11.0 \\
Finnish & 352 & 63 & 7.6 & 7.1 & 10.2 & 2.7 & 2.6 & 3.8 \\
French & 62 & 227 & 9.1 & 10.0 & 8.8 & 5.4 & 6.4 & 5.2 \\
Gujarati & 149 & 283 & 13.3 & 12.7 & 13.7 & 6.2 & 6.2 & 6.2 \\
Hindi & 120 & 119 & 15.1 & 15.5 & 14.7 & 6.6 & 8.3 & 5.1 \\
Kazakh & 136 & 232 & 7.8 & 7.4 & 8.0 & 3.1 & 2.4 & 3.6 \\
Khmer & 98 & 228 & 26.1 & 22.8 & 27.6 & 14.1 & 11.2 & 15.4 \\
Kannada & 245 & 123 & 10.0 & 10.4 & 9.3 & 5.2 & 5.2 & 5.1 \\
Korean & 93 & 133 & 24.3 & 26.0 & 22.9 & 13.5 & 14.0 & 13.1 \\
Kyrgyz & 273 & 149 & 10.6 & 12.1 & 7.9 & 4.5 & 5.7 & 2.1 \\
Ganda & 228 & 78 & 13.4 & 13.3 & 13.6 & 8.2 & 7.9 & 9.0 \\
Latvian & 177 & 179 & 7.0 & 6.6 & 7.3 & 3.0 & 2.7 & 3.3 \\
Marathi & 148 & 295 & 13.2 & 9.3 & 15.0 & 8.2 & 4.8 & 9.9 \\
Polish & 95 & 243 & 6.6 & 7.2 & 6.3 & 3.3 & 3.8 & 3.1 \\
Russian & 203 & 153 & 8.0 & 7.4 & 8.9 & 4.0 & 3.7 & 4.6 \\
Shona & 118 & 275 & 10.5 & 8.6 & 11.4 & 4.1 & 2.2 & 5.1 \\
Swedish & 64 & 266 & 10.1 & 8.5 & 10.5 & 5.6 & 4.6 & 5.9 \\
Swahili & 59 & 152 & 7.7 & 11.1 & 6.5 & 3.7 & 3.9 & 3.6 \\
Tamil & 214 & 163 & 15.3 & 17.0 & 13.2 & 8.9 & 10.1 & 7.4 \\
Telugu & 76 & 235 & 12.7 & 11.3 & 13.1 & 7.9 & 8.2 & 7.8 \\
Tajik & 116 & 123 & 11.1 & 12.5 & 9.9 & 4.3 & 4.9 & 3.7 \\
\midrule
Average & - & - & 12.4 & 12.3 & 12.4 & 6.5 & 6.2 & 6.4 \\
\bottomrule
\end{tabular}
\vspace{1mm}
\caption{
\textbf{Analysis of Gender Bias.}
We compare ASR models trained on \MMS{} data and FLEURS data.
We report dev CER per gender of the speakers for 27 languages of FLEURS for which \MMS{} provides data and for which there are at least 50 samples for each gender. 
}
\label{tab:rai-gender-full-results} 
\end{table}
}

\newcommand{\insertTableASREverything}{
\begin{table}[t]
\centering
\begin{tabular}{lrrrrr}
\toprule
& \#lang & FLEURS & CV & VP & MLS \\
\midrule
\textit{Prior Work} \\
VoxPopuli~\citep{wang2021voxpopuli} & 1 & & & 15.3 \\
Maestro~\citep{chen2022maestro}  & 14 & & & 8.1 \\
RNN-T 1B~\citep{boli2021scaling} & 15 & & & & 7.9 \\
Whisper~\citep{radford2022whisper} & 99 & & & 13.6$^*$ & 7.3* \\
ML-IO~\citep{tjandra2022massively} & 70 & & & & 7.5 \\
USM-M~\citep{zhang2023usm} & 102 & 6.5 \\
\midrule
\textit{This Work - Single-Domain training} \\
FL & 102 & 6.4 & & & \\
CV & 76 & & 19.7 & & \\
VP & 14 & & & 10.3 & \\
MLS & 8 & & & & 8.7 \\
\midrule
\textit{This Work - Multi-Domain training} \\
\MMS{}+FL+CV+VP+MLS & 1,162 & 6.2 & 19.6 & 10.6 & 9.0 \\
\bottomrule
\end{tabular}
\vspace{1mm}
\caption{
\textbf{Evaluation of \modelname{} on Multilingual Benchmarks.} 
\modelname{} is fine-tuned on \MMS{}, FLEURS, CommonVoice, Voxpopuli and MLS. 
We report CER on FLEURS and WER on the other benchmarks.
Results are averaged over all languages of a benchmark and $^{(*)}$ indicates results with a different data normalization which does not enable a strict comparison to other results~\citep{radford2022whisper}.
Our models use n-gram language models trained on Common Crawl during inference.
For each approach, we show the number of languages an individual models supports and some prior results are based on multiple models, e.g., \citet{wang2021voxpopuli}.
} 
\label{tab:asr-everything}
\end{table}
}

\newcommand{\insertResultsASRUSMFull}{
\begin{table}[h!]
\centering
\begin{small}
\begin{minipage}[t]{0.49\textwidth}
\begin{tabular}{lrrrr}
\toprule
& Dev  &  Test   & LM    \\
& CER  &  CER   &     \\
\midrule
Afrikaans & 6.3 & 6.3 & CC LM \\
Amharic & 6.8 & 7.0 & CC LM \\
Arabic & 4.3 & 4.9 & CC LM \\
Assamese & 7.3 & 7.7 & CC LM \\
Asturian & 5.2 & 5.1 & CC LM \\
Azerbaijani & 4.2 & 3.9 & CC LM \\
Belarusian & 3.8 & 3.8 & CC LM \\
Bengali & 5.1 & 5.3 & CC LM \\
Bosnian & 3.8 & 3.4 & FL LM \\
Bulgarian & 3.0 & 3.2 & CC LM \\
Catalan & 2.8 & 2.9 & CC LM \\
Cebuano & 3.6 & 4.4 & CC LM \\
Czech & 3.3 & 3.0 & CC LM \\
Sorani & 6.5 & 7.4 & FL LM \\
Mandarin & 14.8 & 14.9 & CC LM \\
Welsh & 5.7 & 5.9 & CC LM \\
Danish & 5.5 & 5.7 & CC LM \\
German & 3.0 & 2.7 & CC LM \\
Greek & 4.0 & 3.7 & CC LM \\
English & 4.6 & 4.3 & CC LM \\
Estonian & 2.9 & 2.7 & FL LM \\
Persian & 4.0 & 4.1 & FL LM \\
Finnish & 2.6 & 2.4 & FL LM \\
French & 4.1 & 4.1 & CC LM \\
Fula & 13.9 & 13.8 & FL LM \\
Irish & 19.3 & 19.5 & CC LM \\
Galician & 2.8 & 2.7 & CC LM \\
Gujarati & 5.2 & 5.1 & CC LM \\
Hausa & 5.5 & 5.9 & CC LM \\
Hebrew & 15.5 & 12.9 & CC LM \\
Hindi & 5.4 & 4.7 & CC LM \\
Croatian & 3.5 & 3.2 & CC LM \\
Hungarian & 4.1 & 4.2 & CC LM \\
Armenian & 3.2 & 3.2 & FL LM \\
Igbo & 10.7 & 11.1 & FL LM \\
Indonesian & 2.3 & 2.1 & CC LM \\
Icelandic & 5.2 & 4.1 & CC LM \\
Italian & 1.5 & 1.4 & CC LM \\
Javanese & 4.3 & 3.8 & CC LM \\
Japanese & 14.5 & 14.6 & CC LM \\
Kamba & 12.5 & 11.3 & FL LM \\
Kannada & 4.6 & 4.4 & FL LM \\
Georgian & 3.4 & 3.6 & CC LM \\
Kazakh & 2.9 & 3.1 & CC LM \\
Kabuverdianu & 4.3 & 4.3 & FL LM \\
Khmer & 9.9 & 10.3 & CC LM \\
Kyrgyz & 4.1 & 3.6 & FL LM \\
Korean & 11.4 & 11.8 & CC LM \\
Lao & 26.5 & 24.1 & CC LM \\
Latvian & 2.8 & 2.2 & CC LM \\
Lingala & 4.0 & 4.3 & FL LM \\
Lithuanian & 4.3 & 3.7 & CC LM \\
\bottomrule
\end{tabular}
\end{minipage} \hfill
\begin{minipage}[t]{0.49\textwidth}
\begin{tabular}{lrrrr}
\toprule
& Dev  &  Test   & LM    \\
& CER  &  CER   &     \\
\midrule
Luxembourgish & 8.0 & 7.5 & CC LM \\
Ganda & 8.0 & 8.4 & FL LM \\
Luo & 4.9 & 5.0 & FL LM \\
Malayalam & 4.4 & 4.2 & FL LM \\
Marathi & 7.2 & 6.6 & CC LM \\
Macedonian & 2.0 & 1.9 & CC LM \\
Maltese & 3.7 & 3.7 & CC LM \\
Mongolian & 5.6 & 5.9 & CC LM \\
Maori & 5.9 & 6.9 & CC LM \\
Burmese & 8.9 & 9.2 & CC LM \\
Dutch & 3.7 & 3.1 & CC LM \\
Norwegian & 3.7 & 4.1 & CC LM \\
Nepali & 8.3 & 7.7 & CC LM \\
Northern & 6.8 & 6.1 & FL LM \\
Nyanja & 6.3 & 6.7 & CC LM \\
Occitan & 7.8 & 8.2 & CC LM \\
Oromo & 15.7 & 16.2 & CC LM \\
Oriya & 6.3 & 7.1 & CC LM \\
Punjabi & 7.4 & 7.0 & CC LM \\
Polish & 2.6 & 2.7 & CC LM \\
Portuguese & 2.8 & 2.8 & CC LM \\
Pashto & 13.1 & 14.1 & CC LM \\
Romanian & 3.7 & 3.1 & CC LM \\
Russian & 3.1 & 3.0 & CC LM \\
Slovak & 2.7 & 2.5 & CC LM \\
Slovenian & 4.0 & 3.7 & CC LM \\
Shona & 3.8 & 4.1 & FL LM \\
Sindhi & 7.2 & 7.2 & FL LM \\
Somali & 12.8 & 13.0 & CC LM \\
Spanish & 1.8 & 2.1 & CC LM \\
Serbian & 10.3 & 12.7 & FL LM \\
Swedish & 4.7 & 4.6 & CC LM \\
Swahili & 3.3 & 3.4 & CC LM \\
Tamil & 8.2 & 9.0 & FL LM \\
Telugu & 6.6 & 6.9 & CC LM \\
Tajik & 4.0 & 4.5 & CC LM \\
Filipino & 3.1 & 3.1 & CC LM \\
Thai & 7.6 & 8.3 & CC LM \\
Turkish & 3.5 & 3.1 & CC LM \\
Ukrainian & 3.3 & 2.9 & CC LM \\
Umbundu & 10.7 & 10.2 & FL LM \\
Urdu & 9.8 & 8.1 & CC LM \\
Uzbek & 4.7 & 5.0 & CC LM \\
Vietnamese & 5.9 & 6.1 & CC LM \\
Wolof & 11.2 & 11.5 & FL LM \\
Xhosa & 6.0 & 6.1 & FL LM \\
Yoruba & 17.0 & 16.1 & FL LM \\
Cantonese & 12.9 & 12.4 & CC LM \\
Malay & 2.8 & 2.6 & CC LM \\
Zulu & 5.2 & 5.5 & FL LM \\
\midrule 
Average & 6.3 & 6.3 \\
\bottomrule

\end{tabular}
\end{minipage}
\end{small}
\vspace{1mm}
\caption{
\textbf{Results on FLEURS-102.}
We show character error rate on the dev and test sets of all 102 FLEURS languages for \modelname{} when fine-tuned on the labeled data of FLEURS. We use language-specific adapters and heads.
For inference we choose between two n-gram language models (LM) based on dev set accuracy:
a word-based model trained on Common Crawl (CC LM) or a character-based model trained on the FLEURS training transcriptions (FL LM).
} 
\label{tab:asr-usm-perlang-devtest} 
\end{table}
}

\newcommand{\insertTableResultsASRFullEval}{
\begin{table}
\centering
\begin{tabular}{lrrrrr}
\toprule
& \#lang & CER & CER $\leq$ 5 & \% \\
\midrule
Asia & 335 & 1.6 \ci{0.1} & 330 & 99\% \\
South America & 136 & 1.5 \ci{0.2} & 132 & 97\% \\
North America & 144 & 2.2 \ci{0.2} & 139 & 97\% \\
Europe & 41 & 1.7 \ci{0.4} & 40 & 98\% \\
Africa & 363 & 2.9 \ci{0.2} & 331 & 91\% \\
Pacific & 88 & 1.7 \ci{0.5} & 87 & 99\% \\
\midrule
& 1,107 & 2.1 \ci{0.1} & 1,059 & 96\% \\
\bottomrule
\end{tabular}
\vspace{1mm}
\caption{
\textbf{ASR Evaluation on \nummmslang{} Languages.}
We evaluate the \modelname{} multi-domain model trained on \MMS{}, FLEURS, CommonVoice, VoxPopuli, and MLS supporting 1,162 languages (\textsection\ref{sec:asr-everything})
Results are in terms of the average character error rate on the \MMS{} test sets for the languages of different geographical regions.
We also show the number of languages for which the model achieves CER less than five, indicating systems which on average produce no more than one incorrect character every twenty characters.
All results are shown with confidence interval 95\%.
} 
\label{tab:asr-full-eval}
\end{table}
}

\title{Scaling Speech Technology to 1,000+ Languages}
\author{
\centering 
\centerline{
Vineel Pratap$^{\Diamond}$\quad
Andros Tjandra$^{\Diamond}$\quad
Bowen Shi$^{\Diamond}$ 
}\vspace{4mm}
\and
\centerline{\textbf{
Paden Tomasello\quad
Arun Babu\quad
Sayani Kundu\thanks{JPMorgan Chase. Work done while at Meta AI.}\quad
Ali Elkahky\thanks{Apple. Work done while at Meta AI.}
}}\vspace{4mm}\and
\centerline{\textbf{
Zhaoheng Ni\quad
Apoorv Vyas\quad
Maryam Fazel-Zarandi\quad
Alexei Baevski\thanks{Character.AI. Work done while at Meta AI.}
}}\vspace{4mm}\and
\centerline{\textbf{
Yossi Adi\thanks{Meta AI \& Hebrew University of Jerusalem.}\quad
Xiaohui Zhang\quad
Wei-Ning Hsu\quad
Alexis Conneau\thanks{Open AI. Work done while at Meta AI.}
}}\vspace{4mm}\and
\centerline{\textbf{
Michael Auli$^{\Diamond}$
}}
\vspace{4mm}
\and
{\normalfont Meta AI}\and
{\normalfont $^{\Diamond}$ core team}
}
\begin{document}

\maketitle
\vspace*{-0.18cm}
\begin{abstract}
Expanding the language coverage of speech technology has the potential to improve access to information for many more people.
However, current speech technology is restricted to about one hundred languages which is a small fraction of the over 7,000 languages spoken around the world.
The Massively Multilingual Speech (\modelname{}) project increases the number of supported languages by 10-40x, depending on the task.
The main ingredients are a new dataset based on readings of publicly available religious texts and effectively leveraging self-supervised learning.
We built pre-trained wav2vec 2.0 models covering \numptlang{} languages, a single multilingual automatic speech recognition model for \nummmslang{} languages, speech synthesis models for the same number of languages, as well as a language identification model for \numlidlang{} languages.
Experiments show that our multilingual speech recognition model more than halves the word error rate of Whisper on 54 languages of the FLEURS benchmark while being trained on a small fraction of the labeled data.
The \modelname{} models are available at \url{https://github.com/pytorch/fairseq/tree/master/examples/mms}.
\end{abstract}

\section{Introduction}

Speech technology has made much progress over the past decade~\citep{chan2015las,graves2006ctc,baevski2020wav,radford2022whisper} and has been integrated into many consumer products, such as home assistants and smartphones.
Despite this progress, speech technology is still absent for the vast majority of the over 7,000 languages spoken around the world~\citep{lewis2016ethnologue}.
Moreover, many of these languages are at risk of disappearing by the end of this century and the narrow language coverage of current technology may contribute to this trend~\citep{bromham2021disappear}.

Speech models have traditionally been built by training models on large amounts of labeled training data which is only available for a small number of languages.
More recently, self-supervised speech representations have dramatically lowered the amount of labeled data required to build speech systems~\citep{oord2018cpc,schneider2019wav2vec,baevski2020wav}.
But despite this progress, prominent recent work still only supports about 100 languages~\citep{radford2022whisper,zhang2023usm}.

To address this, we build a new dataset comprising a moderate amount of labeled data for \nummmslang{} languages and another dataset of unlabeled speech in \numgrnlang{} languages (\textsection\ref{sec:datasets}). 
We leverage this data to pre-train wav2vec 2.0 models supporting several times more languages than any known prior work (\textsection\ref{sec:pretrain}) and then fine-tune these models to build a multilingual speech recognition model supporting \nummmslang{} languages (\textsection\ref{sec:asr}), language identification models for \numlidlang{} languages (\textsection\ref{sec:lid}) and text-to-speech models for \nummmslang{} languages (\textsection\ref{sec:tts}). 

\begin{figure}[t]
\centering
\includegraphics[width=\textwidth]{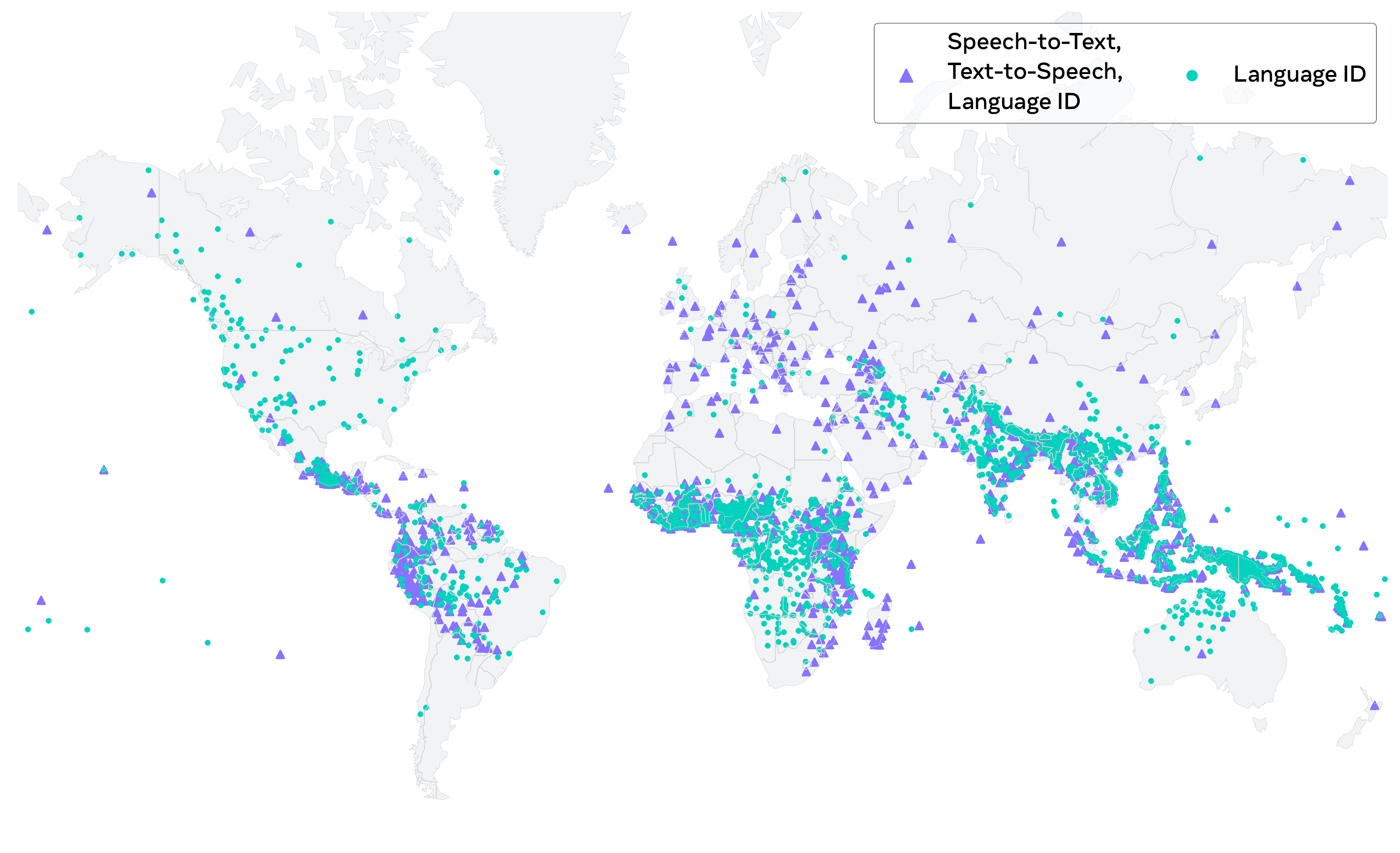} 
\caption{Illustration of where the languages supported by MMS are spoken around the world: 
MMS models support speech-to-text and text-to-speech for \nummmslang{} languages as well as language identification for \numlidlang{} languages.}
\label{fig:map}
\end{figure}

The Massively Multilingual Speech (MMS) project aims to expand speech technology to many more people and we hope that it can be a small contribution to preserving the languages diversity of this world.
\autoref{fig:map} illustrates the location where the languages supported in this work are spoken and which tasks we cover for each language.

\section{Related Work}

\paragraph{Multilingual Speech Datasets.} 
Current multilingual speech datasets with transcriptions are drawn from a variety of domains, including Wikipedia~\citep{ardila2019common,conneau2022fleurs}, political speech~\citep{wang2021voxpopuli}, audiobooks~\citep{pratap2020mls} to name a few.  
They are limited to about 100 languages with some datasets only containing data for European languages \citep{wang2021voxpopuli,pratap2020mls}.
Multilingual datasets without transcriptions include VoxLingua107~\citep{valk2020voxlingua} spanning data in 107 languages, or VoxPopuli~\citep{wang2021voxpopuli} which contains large amounts of unlabeled data for European languages.
\citet{leong2022bloom} covers 56 low-resource languages and 428 hours of data.

Compared to prior work utilizing read versions of the New Testament~\cite{black2019cmu}, MMS covers many more languages (58\% more for \MMS{} and over five times more for \GRN{}) and our alignments are of much higher quality which leads to better models as we show in~\textsection\ref{sec:data-comp-cmu}.
We also use the resulting data to train self-supervised models, build speech recognition systems as well as language identification models while as~\citet{black2019cmu} focused on speech synthesis.

\paragraph{Multilingual Automatic Speech Recognition (ASR).}
Prior work on multilingual speech recognition includes both non-neural methods~\citep{burget2010multilingual,lin2009study}, hybrid neural and HMM models~\citep{heigold2013multilingual} and more recently neural systems~\citep{cho2018multilingual,toshniwal2018multilingual,kannan2019mult,li2019bytes}.
\citet{boli2021scaling} built multilingual ASR for up to 15 languages, \citet{Pratap2020multiasr,lugosch2022icassp, tjandra2022massively} trained and explored several strategies for 50\texttt{+} languages. 

Whisper~\citep{radford2022whisper} mines 680K hours of data from the web and their model supports the transcription of 99 different languages. 
\cite{zhang2023usm} trained a multilingual ASR model based on YouTube audio data and successfully scaled to 100 languages. 
A notable exception is~\citet{li2022asr2k} who created ASR for 1,909 languages by mapping the phoneme-like output of an eight language multilingual model to appropriate phonemes for the language of interest. 
In contrast, MMS uses actual paired speech and text for over 1,100 languages and present a comparison to their approach below (\textsection\ref{sec:data-comp-asr2k}).

\paragraph{Spoken Language Identification (LID).}
Whisper~\citep{radford2022whisper} also supports language identification and can distinguish between 99 different languages. 
\citet{fan2021exploring} utilizes wav2vec 2.0 for language identification using the API17-OLR dataset that consists of ten Asian languages. 
Later, \citet{tjandra2022improved} demonstrated that a cross-lingual self-supervised model can improve language identification performance by training a language identification model for 26 languages using a proprietary dataset. 
\citet{babu22_interspeech} fine-tunes a pre-trained model to perform LID for 107 languages using the VoxLingua-107 dataset~\citep{valk2020voxlingua}.  
In this work, we scale the number of languages to over 4,000 which to our knowledge is the broadest coverage spoken language identification model so far.

\paragraph{Multilingual Text-To-Speech (TTS).} 
Speech synthesis has been undergoing a transition from controlled settings to the generation of more diverse speech, with multilinguality being a crucial aspect~\citep{casanova2022yourtts,zhang2023speak}. 
However, the lack of multilingual training data, particularly for low-resource languages, presents a common obstacle in scaling TTS to more languages.
To address data scarcity, prior work explored various approaches including byte encoding to unify text representations which was evaluated on English, Spanish, and Chinese~\citep{li2019bytes}
Further studies explored other input representations, including phonemes~\citep{zhang2019learning} and phonological features \citep{Staib2020Phonological}. 

Additionally, various modeling schemes have been developed to encourage knowledge sharing between languages, such as parameter generation networks~\citep{Nekvinda2020one} and leveraging unpaired speech or text data for pre-training~\citep{Saeki2023Virtuoso,Saeki2023learning}.
Despite these efforts, most prior work still covers a small number of languages but there are a few efforts which scaled to 46 languages~\citep{he2021multi} or use unsupervised techniques to scale to 101 languages~\citep{Saeki2023Virtuoso}. 
\citet{meyer2022bibletts} also builds VITS models based on readings of the Bible but their work is limited to ten African languages.

\paragraph{Multilingual NLP.}
Multilinguality has been a very active research area in NLP where researchers introduced cross-lingually pre-trained sentence encoders~\citep{conneau2019cross} spanning 100 languages, or pre-trained multilingual sequence to sequence models~\citep{liu2020multilingual} applied to machine translation. 
Multilingual machine translation models have also been scaled to 200 languages~\citep{nllbteam2022language} and even 1,000 languages~\citep{bapna2022mt1k}.

\section{Dataset Creation}
\label{sec:datasets}

Our work leverages two new datasets to expand the language coverage of speech technology.
In this section, we first detail how we create a labeled dataset which includes speech audio paired with corresponding text in \nummmslang{} languages (\MMS{}; \nummmshrs{} hours; \textsection\ref{sec:data-paired}). 
Second, we discuss the creation of an unlabeled dataset for which we only have audio recordings and no corresponding text.
This dataset spans \numgrnlang{} languages (\GRN{}; {\numgrnhrs} total hours; \textsection\ref{sec:data-unpaired}). 

We also use an unlabeled version of \MMS{} for pre-training and language identification. 
This spans a larger number of languages, as we can also use unlabeled audio from our data source (\MMSU{}; \nummmsulang{} languages; {\nummmsuhrs} hours). 
\autoref{fig:datasets-comp} compares the datasets to existing corpora.
A full list of the languages supported is available at \url{https://github.com/facebookresearch/fairseq/tree/main/examples/mms}.

\insertDatasetLanguageComparison

\subsection{Paired Data for \nummmslang{} Languages (\MMS{})}
\label{sec:data-paired}

We obtain speech data and transcriptions for \nummmslang{} languages by aligning New Testament texts obtained from online sources (\textsection\ref{sec:data-paired-source}) using the following steps:

\begin{enumerate}
    \item Download and preprocess both the speech audio and the text data (\textsection\ref{sec:data-preprocess}). 
    
    \item Apply a scalable alignment algorithm which can force align very long audio files with text and do this for data in 1000+ languages (\textsection\ref{sec:data-align-algo}).

    \item Initial Data Alignment: we train an initial alignment model using existing multilingual speech datasets covering 8K hours of data in 127 languages and use this model to align data for all languages 
    (\textsection\ref{sec:data-initial-align}).  \label{list:flowchart_step3}
    
    \item Improved Data Alignment: we train a second alignment model on the newly aligned data for which the original alignment model has high confidence and generate the alignments again. 
    The new alignment model supports 1,130 languages and 31K hours of data including the data used in step~\ref{list:flowchart_step3} (\textsection\ref{sec:data-align2}).

    \item Final data filtering: we filter the low-quality samples of each language based on a cross-validation procedure. 
    For each language, we train a monolingual ASR model on half of the aligned data to transcribe the other half of the data.  
    We retain only samples for which the transcriptions are of acceptable quality (\textsection\ref{sec:data-cross-val}).
    
    \item We partition the data into training, development and test portions (\textsection\ref{sec:data-split}).
\end{enumerate}

\subsubsection{Data Source}
\label{sec:data-paired-source}

The \MMS{} dataset is based on recordings of people reading the New Testament in different languages.
The New Testament consists of 27 books and a total of 260 chapters. 
Specifically, we obtain data from Faith Comes By Hearing\footnote{\url{https://www.faithcomesbyhearing.com/}}, goto.bible and bible.com. 
This includes the original text data as we well as the corresponding audio recording.

\paragraph{Basic Data Characteristics.}
The data sources provide 1,626 audio recordings of the New Testament in \nummmsulang{} languages, totaling \nummmsuhrs{} hours and we refer to this data as the \MMSU{} dataset.\footnote{For the audio files with no paired text, we perform VAD and segment them into smaller files.}
Out of these, both text and audio is available for 1,306 different recordings in 1,130 languages and a total of 49K hours, which we focus on for \MMS{}.
For 99 languages, we have multiple recordings.
Each recording provides separate audio files for each chapter and the duration of each chapter is on average 6.7 minutes but there is significant variance, depending on the language and chapter.
Recordings are almost always single speaker which makes it well suited for building speech synthesis systems (\textsection\ref{sec:tts}).
However, speakers are often male which may introduce unwanted biases into machine learning models which we analyze below (\textsection\ref{sec:rai-gender-bias}).

\paragraph{Multiple Scripts or Dialects per Language.}
When there are multiple recordings per language, we found that some recordings differ in the used writing script, e.g., for Serbian there are recordings using the Latin script and another which uses Cyrillic.\footnote{Specifically, we measure the correlation of the character frequency distribution for all pairs of the recordings and examine recordings with a correlation lower than 0.99 further.}
Recordings can also differ in the spoken dialect, e.g., for the Poqomchi' language there are recordings with western and eastern dialects. 

Depending on the downstream task, we handle these cases differently: for language identification, we merge the different recordings regardless of writing script or dialect, for speech synthesis models, we choose one recording per dialect/script to avoid introducing additional speakers into the training data, and for automatic speech recognition, we combine all the recordings within the same script/dialect into one and treat them as different languages, e.g., srp-script:latin, srp-script:cyrillic.

\paragraph{Micro and Macro Language Distinction.}
When there is a micro and macro language distinction, then we generally keep micro languages as distinct languages. 
However, for languages which are part of benchmarks we use for evaluation, e.g., FLEURS~\citep{conneau2022fleurs}, we deviate from this policy if the respective benchmark only contains the macro language, by merging the micro languages.
For example, Azerbaijani is a macro language and \MMS{} provides data for two associated micro languages, North Azerbaijani and South Azerbaijani.
For ASR, our evaluation benchmarks also keep the same distinction and so we model both micro languages separately, however, for LID, one of our benchmarks, VoxLingua-107~\citep{valk2020voxlingua}, only contains the macro language, and therefore, for LID only, we merge both micro languages into a single macro language.

\paragraph{Recordings with Background Music.}
Some of the recordings contain background music and we refer to these as drama recordings.
In our final \MMS{} dataset, 38\% of languages are represented solely by a drama recording and 11\% have both drama recordings and recordings without background music (non-drama recordings).
For speech synthesis, we apply pre-processing to remove background music (\textsection\ref{sec:tts}).

While this data source covers a lot of languages, it requires careful curation to make it usable for building high-quality models which presents unique challenges given the large number of languages. 
We detail the steps we take in the remainder of this section.

\subsubsection{Data Pre-processing}
\label{sec:data-preprocess}

\paragraph{Speech.}
The original audio files are available in MP3 stereo format using a 22/24/44 kHz sampling rate. 
We convert all of them to a single channel and 16 kHz sampling rate. 

\paragraph{Text Normalization.}
We design a generic text normalization pipeline that works well across the languages we consider. 
First, we perform NFKC normalization and lower case all characters.\footnote{
For example, there are two ways to represent Ç (Latin C with combining cedilla): splitting it into Latin capital C and combining cedilla (NFKD), or having a single Unicode character C with cedilla.} 
NFKC normalization helps to make sure the character encoding is consistent.
Next, we remove HTML tags such as "\&gt;" or "nbsp;".
We also remove punctuation and try to perform this carefully by including characters for which we are confident that they are in fact punctuation.\footnote{We obtain an initial set of punctuation characters from the respective Unicode category.}

We noticed that some recordings have a relatively high rate of brackets in the text: 
our criteria is recordings where at least 3\% of verses contain brackets which resulted in about 50 recordings. 
We listened to a few instances of each recording to verify whether the text in the brackets is present in the audio or not. 
In many cases, we noticed that the text in the brackets was not spoken and we removed the brackets and the text within.

\subsubsection{Scalable Forced Alignment}
\label{sec:data-align-algo}

The chapter recordings from the data source can be up to 43 minutes long which cannot be directly used by current machine learning algorithms:
we use Transformer models which require large amounts of GPU memory and their computational complexity is quadratic in the input size.
We therefore segment the data into smaller units so it can be used by standard algorithms.
Forced alignment determines which parts of the audio correspond to which parts of the text.

\begin{figure}[t]
\centering
\includegraphics[width=0.9\textwidth]{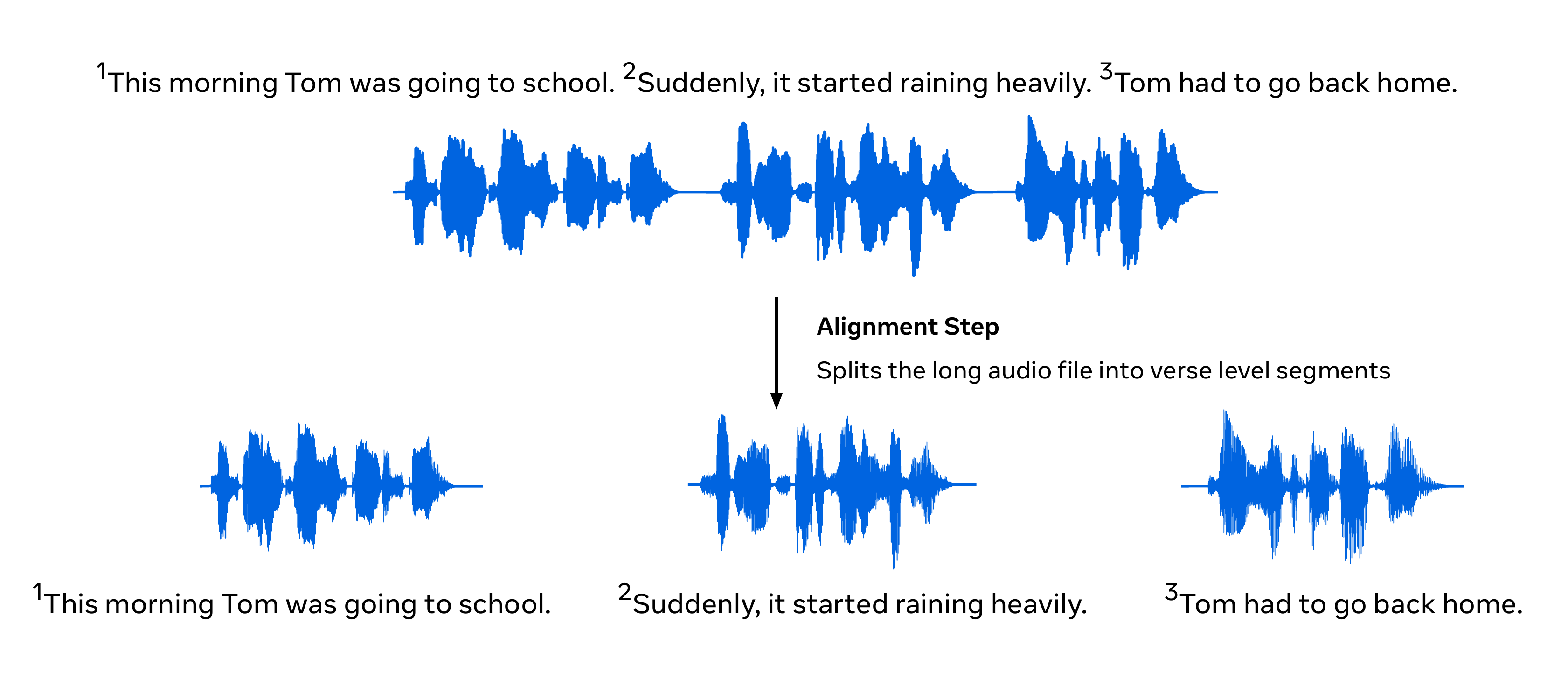}
\caption{
\textbf{Illustration of Data Alignment.}
Forced alignment enables segmenting audio recordings and corresponding text into smaller segments that can be used to train machine learning models.}
\label{fig:align}
\end{figure}

Given this alignment, the data can be segmented in different ways.
In our case, we segment the data into individual verses which are typically a single sentence but can sometimes contain several sentences. 
The average duration of a verse is about 12 seconds.
\autoref{fig:align} illustrates how the alignments enable creating verse-level audio segments for each chapter recording. 
In this section we detail how we perform efficient forced alignment on GPUs.

\paragraph{Generating Posterior Probabilities.}
Forced alignment requires posterior probabilities from an acoustic model which we use for alignment (\textsection\ref{sec:data-initial-align}). 
This acoustic model is a Transformer which requires substantial amounts of memory to store activations which makes it infeasible to use for long audio files.
As a workaround, we chunk the audio files into 15 second segments, generate posterior probabilities for each audio frame using the alignment model, and then concatenate these posterior probabilities into a single matrix again.
The acoustic model is trained with Connectionist Temporal Classification (CTC;~\citealt{graves2006ctc}).

\paragraph{Forced Alignment using CTC.}
Next, we perform forced alignment which finds the most likely path in the posterior probabilities for a given input audio sequence of length $T$ and a text transcription of length $L$.
These posterior probabilities require $\mathcal{O}(T \times L)$ memory and a path will be of length $T$.
This path is computed using the Viterbi algorithm. 
There are open source libraries implementing the algorithm on CPU~\citep{ctcsegmentation,kahn2022flashlight}, however, the CPU versions are slow to run, particularly on long recordings, as we will show below.

\paragraph{Efficient Forced Alignment on GPUs.}
In order to make force alignment efficient for our purpose, we implemented a GPU version that computes the Viterbi path memory in a memory efficient way.
Storing all $\mathcal{O}(T \times L)$ forward values for the Viterbi algorithm is infeasible on GPUs due to memory constraints.
We therefore only store forward values for the current and the previous time-step and regularly transfer the computed backtracking matrices to CPU memory.
This reduces the required GPU memory to $\mathcal{O}(L)$ compared to $\mathcal{O}(T \times L)$ and enables forced alignment for very long audio sequences at high speed.
Appendix~\ref{app:align} illustrates the algorithm and an implementation is available as part of TorchAudio~\citep{yang2021torchaudio}.\footnote{\url{https://github.com/pytorch/audio}} 

\insertForceAlignBenchmark

\autoref{fig:results_fa_benchamrk} shows that the forced alignment implementation scales much better to longer sequences than CPU alternatives such as ctc-segmentation~\citep{ctcsegmentation}, a popular segmentation library used in ESPNet~\citep{watanabe2018espnet}, SpeechBrain~\citep{speechbrain} and Flashlight~\citep{kahn2022flashlight}.

\paragraph{Robust Alignment for Noisy Transcripts.}
For many recordings, speakers introduce the chapter name and the version of the New Testament before reading the first verse, however, the corresponding text does not contain this information.
This is problematic for forced alignment because the algorithm will still try to align the beginning of the audio to the text which can result in incorrect alignments.
Another challenge is that numbers are generally spelled as digits in the text whereas our alignment model is trained on existing corpora which follow common practice of spelling numbers out fully (\textsection\ref{sec:data-initial-align}).
Spelling numbers out requires language-specific and hand-crafted tooling which is not available for the \nummmslang{} languages we consider.

\insertStarPlots 

To enable robust alignment in both cases, we introduce a star token ($\Star$;~\citealt{pratap2022star, cai2022wctc}) to which audio segments can be mapped if there is no good alternative in the text.\footnote{This is different from an OOV token or silence token in an HMM topology, which model a certain type of acoustic behavior and are independent from other in-vocabulary tokens.}
We insert $\Star$ at the beginning of the text data of each chapter and replace numerical digits with $\Star$ throughout. 
The posterior probability for this token is set to one.
After alignment, we add back the original digits and the subsequent data filtering often removes segments where the audio contains additional information not present in the aligned text (\textsection\ref{sec:data-cross-val}).
\autoref{fig:star_token} illustrates how adding the $\Star$ token helps to improve alignment when the paired text does not cover the beginning of the audio.

\subsubsection{Initial Data Alignment}
\label{sec:data-initial-align}

\paragraph{Acoustic Model.}
To perform the forced alignment, prior work typically uses acoustic models trained on data in the same language.
For example, for the eight languages of the MLS dataset~\cite{pratap2020mls} the authors used acoustic models trained on existing data for these languages.
However, our setting includes many languages for which no datasets or acoustic models exist.
We therefore train a multilingual acoustic model on FLEURS~\citep{conneau2022fleurs} and CommonVoice 8.0~\citep{ardila2019common} to learn a shared representation across languages which we hope to generalize to unseen languages. 
The multilingual model is based on fine-tuning XLS-R~\citep{babu22_interspeech} using a total 8K hours of data covering 127 languages.

\insertUromanXsampaExamples

\paragraph{Text Encoding.}
The text data is represented using the uroman transliteration tool~\citep{hermjakob-etal-2018-box}
which maps different writing scripts to a common Latin script representation.\footnote{\url{https://www.isi.edu/~ulf/uroman.html}}
This is done using character descriptions based on Unicode tables and a large number of additional heuristics and it has language-specific romanization rules for some languages.
Prior work~\citep{black2019cmu} used Unitran~\citep{yoon-etal-2007-multilingual,qian-etal-2010-python,black2019cmu} which converts UTF-8 encoded text into a phonetic transcription in either WorldBet~\citep{hieronymus1993ascii} or X-SAMPA~\citep{wells_1995}. 

Inspired by~\citet{black2019cmu}, we initially investigated X-SAMPA but found that uroman led to similar quality results and we decided to use uroman since it is easier to interpret compared to International Phonetic Alphabet (IPA) based symbols generated by Unitran. 
\autoref{tab:data-ex-uroman-xsampa} shows some example outputs from uroman. 
We lowercase all the letters of the uroman output and retain only \textit{a} to \textit{z} characters as well as the apostrophe character to train acoustic models for forced alignment (\autoref{sec:data-align-algo}).

% \ma{TODO [low pri]: show experiment demonstrating that our 127-lang multilingual alignment model can generalize to unseen languages. 
% Train the same model on ~120 languages, perform initial alignment step, train ASR models and compare to ASR models trained on data aligned by 127-lang alignment model.}

\subsubsection{Improved Data Alignment}
\label{sec:data-align2}

To improve the alignments, we use a subset of good-quality samples to train a new alignment model.
Samples are selected based on a score which is the length-normalized difference between the probability of the forced alignment path $P(Y^{aligned} \mid X)$ and the probability of greedy decoding from the alignment model $P(Y^{greedy} \mid X)$.  
The former is constrained by the text used in the alignment and the latter is unconstrained. 
A large difference between these two quantities may indicate that the alignment is incorrect.\footnote{It may also indicate that the alignment model is of poor quality but in practice we found that the metric identifies many incorrect alignments.}
Concretely, the score is 
\begin{equation}
    \centering
      \frac{1}{T} \log P(Y^{aligned} \mid X) - \log P(Y^{greedy} \mid X)
\end{equation}
where $T$ is the length of the audio.
The score can vary from $-\inf$, indicating low quality samples, to $0$, indicating high quality samples. 
After manual inspection of sample quality and their corresponding score for several languages, we select $-0.2$ as the threshold and choose samples with scores greater than this threshold.
The improved alignment model is trained on a total of 31K hours in 1,130 languages which includes the data we used for the initial alignment model.\footnote{This model is available at \url{https://github.com/facebookresearch/fairseq/tree/main/examples/mms}}
We use this new model to re-generate verse level alignments for our data.

\subsubsection{Final Data Filtering}
\label{sec:data-cross-val}

After generating the improved alignments, we noticed that some samples are still of low quality.
Some recordings are not entirely faithful to the text and speakers sometimes add their own interpretation or paraphrase parts of the text. 
This will negatively impact ASR and TTS models trained on the data and we therefore perform a final data filtering step to improve data quality as much as possible.

We we train monolingual ASR models on half of the the aligned samples of each recording, measure performance on the remaining half and remove samples which have a character error rate (CER) in excess of 10\%. 
This removes about 1.7\% of all samples across all languages. 

High CER may be due to low quality samples in either half of the data, that is the training data of the ASR model or the data being evaluated. 
To help ensure we use recordings that are generally of good quality, we remove 3837 recordings which have CER in excess of 5\% on the developmet set.
We retain a total of 1,239 recordings covering 1,107 languages.

\subsubsection{Creating train/dev/test splits}
\label{sec:data-split}

\insertMMSHours

The recordings often contain only a single speaker which makes it challenging to measure generalization performance of models trained on this data.
This is why we evaluate models trained on \MMS{} data on existing benchmarks as much as possible in the remainder of this paper.
The advantage of this is that it side steps the aforementioned issues and it also enables us to better understand how the data can be useful in other domains.

However, existing benchmarks only cover a small fraction of the languages in \MMS{} and in order to be able to develop models for all languages, we split the aligned recordings into train/dev/test splits.
We aim to make the content of the splits as disjoint from each other as possible and use a similar split across different recordings.
To do so, we try to use different books for each split and use the same books for each split across recordings and languages as much as possible.

Concretely, we use the book Mark (MRK) as development set, the book John (JHN) as test set and the remaining books for training. 
For the 147 recordings where not all 260 chapters are available, we deviate from this and make a best effort split by books depending on which books are available in the respective recording.
In this case, we aim to have at least 10\% of all available data in the development set and the test set each, or at most two hours of data in each set, whichever is less.

The final dataset contains \nummmshrs{} hours of paired speech data where we use 36.8K hours for training (82.3\%), 3.5K hours for development (7.8\%) and 4.4K hours for testing (9.9\%).
For each language, the train split contains an average of 32 hours (stddev=19), the dev split contains an average of 3.1 hours (stddev=1.8) and test split an average of 3.9 hours (stddev=2.3).
\autoref{fig:data-mmshrs} shows the data distribution across languages.

\subsection{Unpaired Data for \numgrnlang{} Languages (\GRN)}
\label{sec:data-unpaired}

\paragraph{Data Source.}
The data source for this dataset is Global Recordings Network which provides recordings of Bible stories, evangelistic messages, scripture readings, and songs in more than 6,255 languages and dialects.\footnote{\url{https://globalrecordings.net/}}
The audio files are not accompanied by a corresponding text transcription but the source makes clear which language is being spoken.
We group the data by language, combining  dialects of the same language resulting in a total of 3,860 languages and 9,345 hours of audio.

\insertGRNHours

\paragraph{Pre-processing.}

We convert the audio files to single channel and a sample rate of 16kHz. 
Next, we use inaSpeechSegmenter~\citep{ddoukhanmirex2018}, a CNN-based audio segmentation model, to  identify segments of speech, music, noise and silence in the audio. 
If two segments of speech are separated by intermediate segments containing music or noise, then we consider joining these segments if the intermediate segment is no longer than 20\% of all segments together. 
This is to build samples that are of longer duration and still contain mostly speech.
The remaining non-speech segments are discarded.

Next, we randomly split the speech segments into portions of between 5.5 and 30 seconds. 
This makes the data usable for training downstream models and creates a length distribution of the samples that is in line with other datasets such as FLEURS where the average sample length is about 12 seconds.

\paragraph{Dataset Split.}
Finally, we split the samples of each language randomly into 80\% training data, 10\% development data, and 10\% test data. 
We also remove 51 languages for which we have less than 5 minutes of training data to ensure we have sufficient data to train models.
The final dataset comprises a total of \numgrnhrs{} hours of data in \numgrnlang{} languages.
The training portion is 6.2K hours and there are 770 hours for the valid and test sets each.
For each language, the train set contains an average of 97 minutes (stddev=177.4), and the dev/test sets contain an average of 12.1 minutes (stddev=22.3).
\autoref{fig:data-grnhrs} shows the data distribution across languages.

\subsection{Comparison to Existing Broad Coverage Approaches and Other Datasets}

In this section, we present a comparison to two related studies which aimed to expand speech technology to many languages.
The CMU Wilderness project~\citep{black2019cmu} also used New Testament data to build speech synthesis models for 699 languages (\textsection\ref{sec:data-comp-cmu}) and ASR-2K~\citep{li2022asr2k} focused on automatic speech recognition for nearly two thousand languages (\textsection\ref{sec:data-comp-asr2k}).
Finally, we assess the viability of the new data for building machine learning models by comparing the performance of ASR models trained on \MMS{} to models trained on an existing dataset in an out-of-domain setting (\textsection\ref{sec:data-compdata}).

\subsubsection{CMU Wilderness Dataset}
\label{sec:data-comp-cmu}

The most comparable prior work is the CMU Wilderness project which used data from similar sources~\citep{black2019cmu}. 
To better understand how our data creation process compares to their method, we conduct a best effort side-by-side reproduction of their process and use the resulting data to train monolingual ASR models under the same settings.\footnote{We reproduced the data by following the steps outlined at \url{https://github.com/festvox/datasets-CMU_Wilderness}}

To compare the effectiveness of their data creation method to ours (\textsection\ref{sec:data-paired}), we take the original data from our data source and apply either our protocol or the protocol of \citet{black2019cmu}.
For languages where multiple recordings exist, we only use the recordings used in the CMU Wilderness dataset to enable a better comparison.
Next, we use the resulting data to fine-tune XLS-R models~\citep{babu22_interspeech} for monolingual ASR and then measure accuracy in terms of character error rate on the FLEURS dev set.

\autoref{fig:data-compare-cmu} shows that the \MMS{} data preparation process results in better quality ASR models compared to CMU Wilderness with improvements between 2.1\%-4.7\% CER, depending on the language.
Our alignment procedure also retains a much larger amount of the training data compared to the CMU Wilderness protocol:
for Telugu, there are 26.5 hours of data, \MMS{} retains 26.2 hours compared to 11.1 hours for the CMU Wilderness process.
For English, we start with 17.3 hours, \MMS{} retains 17 hours vs. 10.6 hours for CMU Wilderness. 

\insertComparisonWithOtherDatasets

\subsubsection{ASR-2K}
\label{sec:data-comp-asr2k}

The most comparable multilingual ASR work to ours is ASR-2K~\citep{li2022asr2k} which covers 1,909 languages. 
Their approach is based on mapping the output of an eight language multilingual model to appropriate phonemes for the language of interest. 
In contrast, \MMS{} has actual paired speech and text data.

ASR-2K reports average character error rate (CER) on 34 languages of Common Voice 6.0 and uses a language model for decoding. 
\MMS{} covers 22 out of these 34 languages and we use it to train monolingual ASR models without language models.
The monolingual models trained on {\MMS} dataset obtain an average CER of 9.6 on 22 languages.
ASR-2K reports CER 50.9 on 34 languages.
While not a like for like comparison, this difference suggest that \MMS{} enables higher quality ASR models.
We stress that this is a best effort comparison and does not enable strong conclusions.\footnote{We could not obtain a per-language break down of the ASR-2K results.}

\subsubsection{Other Existing Datasets}
\label{sec:data-compdata}

The \MMS{} covers a large number of languages but it also has potential downsides:
it is both from a particular narrow domain and most recordings are from a single speaker. 
This may lead to poor performance on other domains or when the systems are applied to unseen speakers.

To get a better sense of both issues, we train monolingual ASR models by finetuning XLSR~\citep{babu22_interspeech} on \MMS{} and evaluate these models on the FLEURS benchmark.
For comparison, we train another set of ASR models on labeled data from Common Voice which is an existing dataset and does not have the aforementioned downsides: the domain is general and the data contains multiple speakers. 
We control the amount of training data of both datasets by using exactly ten hours of training data and one hour of development data from both datasets.

\autoref{fig:compare-cv} shows that models trained on CommonVoice perform better on 18 languages of FLEURS (average CER 9.3 vs. 12.2) but the models \MMS{} still enable good performance.\footnote{Spanish (spa), Catalan (cat), Polish (pol), German (deu), Portoguese (por), Russian (rus), Ukrainian (ukr), Turkish (tur), Swahili (swh), French (fra), Dutch (nld), Farsi (fas), English (eng), Welsh (cym), Luganda (lug), Arabic (ara), Tamil (tam), Thai (tha).}
This is despite the fact that \MMS{} utterances are often from a single speaker and are from a very narrow domain.
While there is certainly higher quality data for head languages, this result suggests that the quality of the \MMS{} data can enable high quality speech systems for a large number of other languages.

\section{Cross-lingual Self-supervised Speech Representation Learning}
\label{sec:pretrain}

As a first step, we train a self-supervised model of speech representations on the data outlined above as well as other existing public corpora. 
We use wav2vec 2.0 for pre-training on unlabeled data which we later use as the basis for several downstream speech tasks (\citealt{baevski2020wav};~\textsection\ref{sec:pretrain-w2v}).
The resulting models were pre-trained on \numptlang{} languages which is over four times the number of languages of known prior work~(\citealt{zhang2023usm};~\textsection\ref{sec:pretrain-setup}).
The increased language coverage results in better performance for both ASR and LID compared to XLS-R (\citealt{babu22_interspeech};~\textsection\ref{sec:pretrain-results}) which covered 128 languages and is publicly available.

\subsection{Method: wav2vec 2.0 and XLS-R}
\label{sec:pretrain-w2v}

Our work builds on~\citet{babu22_interspeech} who pretrain wav2vec models on data from multiple languages.
The wav2vec project created a series of models for learning self-supervised speech representations~\citep{schneider2019wav2vec,baevski2019vqwav2vec,baevski2020wav} from unlabeled speech data.
The resulting models can then be used to solve downstream speech tasks by fine-tuning them on labeled data or by tackling these tasks without labeled data using unsupervised learning~\citep{baevski2021unsupervised,liu2022endtoend}.

The most prominent one is wav2vec 2.0~\citep{baevski2020wav} which enables building speech recognition models with only ten minutes of labeled data and even no labeled data at all~\citep{baevski2021unsupervised}.
The basic architecture of wav2vec 2.0 is as follows: a convolutional feature encoder $f: \Inp \mapsto \Feat$ maps raw audio~$\Inp$ to latent speech representations $\ze_1, \dots, \ze_T$ which are input to a Transformer $g: \Feat \mapsto \Context$ to output context representations $\cc_1, \dots, \cc_T$~\citep{baevski2019vqwav2vec}.
Each $\ze_t$ represents 25ms of audio strided by 20ms and the Transformer architecture follows BERT~\citep{vaswani2017transformer,devlin2018bert}.

During training the feature encoder representations are discretized to $\zq_1, \dots, \zq_T$ with a quantization module $\Feat \mapsto \QFeat$ to represent the targets in the objective.
The quantization module uses a Gumbel softmax to choose entries form the codebooks and the chosen entries are concatenated~\citep{jegou2011ieee,jang2016gumbel,baevski2019vqwav2vec}.

The model is trained by solving a contrastive task over masked feature encoder outputs.
At training time, spans of ten time steps with random starting indices are masked.
The objective requires identifying the true quantized latent $\zq_t$ for a masked time-step within a set of $K=100$ distractors sampled from other masked time steps.
The objective is augmented by a codebook diversity penalty to encourage the model to use all codebook entries~\citep{dieleman2018challenge}.

XLSR and XLS-R train wav2vec 2.0 on many different languages from several datasets to obtain cross-lingual representations~\citep{conneau2020unsupervised,babu22_interspeech}.
In order to balance the training data, two data sampling steps are performed.
First, for each dataset, we sample the data for the different languages $L$ from a distribution $p_l\sim\left(\frac{n_l}{N}\right)^{\beta_L}$ where $l=1,\ldots,L$, $n_l$ is the amount of unlabeled data for each language in the dataset, $N$ is the total amount of training in the dataset, and $\beta_L$ is the upsampling factor which controls the trade-off between high- and low-resource languages during pretraining.
Second, we balance the different datasets by treating each dataset as a language in the above sampling scheme with a sampling parameter $\beta_D$.

\insertPretrainSetup

\subsection{Pre-training Setup}
\label{sec:pretrain-setup}

\paragraph{Hyperparameters.}
We largely follow prior work in training cross-lingual wav2vec 2.0 models~\citep{conneau2020unsupervised,babu22_interspeech} and use the wav2vec 2.0 implementation available in fairseq~\citep{ott2019fairseq} to train models with roughly 300M and 1B parameters (\autoref{tab:pretrain-models}).
To make efficient use of GPU memory, we use a fully sharded backend~\citep{rajbhandari2021zero} as well as activation checkpointing~\citep{chen20216checkpointing} implemented in FairScale~\citep{FairScale2021}. 

Our models are optimized with Adam~\citep{kingma2015adam} and the learning rate is warmed up for the first 32K steps followed by polynomial decay to zero for the remainder of training. 
Training audio sequences are cropped to a maximum of 320K samples, or 20 seconds, and all models were pre-trained for a total of one million updates on A100 GPUs with 80GB of memory.
The \modelnameb{0.3} model was trained with an effective batch size of 2.3 hours of data across 48 GPUs and the \modelnameb{1} model was trained with an effective batch size of 3.5 hours on 64 GPUs.

\paragraph{Data.}
The pre-training data covers about 491K hours in \numptlang{} languages. 
This data is drawn from six training corpora with different characteristics, including the corpora used in XLS-R~\citep{babu22_interspeech}:

\begin{itemize}
    \item \MMSU: \nummmsulang{} languages comprising \nummmsuhrs{} hours (\textsection\ref{sec:data-paired}). 
    \item Multilingual Librispech (MLS): 8 European languages of read books totaling 50K hours~\citep{pratap2020mls}
    \item CommonVoice (CV): 89 languages totaling 8.8 hours of read Wikipedia text; we use v9.0 of the corpus~\citep{ardila2019common})
    \item VoxLingua-107 (VL): 107 languages totaling 5.3K hours of YouTube content~\citep{valk2020voxlingua}
    \item BABEL (BBL): 17 African and Asian languages totaling about 1K hours of conversational telephone data~\citep{gales2014babel}
    \item VoxPopuli (VP): 371K hours of unlabeled speech data in 23 languages derived from European Parliament event recordings~\citep{wang2021voxpopuli}

\end{itemize}

Pre-training only uses the speech audio and none of the transcriptions and we balance the data following the strategy outlined in~\textsection\ref{sec:pretrain-w2v} using $\beta_L = \beta_D = 0.5$

\subsection{Comparison to XLS-R}
\label{sec:pretrain-results}

\insertPretrainComparisonASR

To better understand how the {\modelname} models compare to XLS-R we fine-tuned both for automatic speech recognition on the 61 languages of the FLEURS benchmark for which \MMS{} provides training data.
Models are evaluated without a language model and we report the average character error rate (CER) on FLEURS development data over all languages.

\autoref{fig:pretrain-comparison-asr} shows that the {\modelname} models perform better: for the 300B size, {\modelname} has 0.6 lower CER than XLS-R, and for the 1B size, the difference is 0.7 CER.
More capacity helps to improve performance: 
for XLS-R the error rate decreases by 3.2 CER absolute when scaling the number of parameters from 300M to 1B and for \MMS{} there is a 3.0 CER improvement.

\modelname{} pre-trains on over ten times the number of languages of XLS-R and this improves performance, particularly on low resource languages (\autoref{fig:pretrain-xlsr-mms-perlang}) such as Amharic (amh), Lao (lao) or Malayalam (mal).
Compared to XLS-R, the pre-training data of \modelname{} covers the following languages which improve: Chewa (nya), Fulah (ful) and Oromo (orm).
However, improvements at low resource languages result in a small degradation in some of the high-resource languages such as English (eng) or Spanish (spa) but there are also other languages such as Tajik (tgjk) or Welsh (cym) which perform less well.

\insertPretainXLSRvsMMS

Equipped with this new pre-trained model we now investigate the use of \MMS{} and \GRN{} for several downstream tasks.

\section{Automatic Speech Recognition}
\label{sec:asr}

We first turn to the task of transcribing speech in up to \nummmslang{} different languages.
We use the labeled dataset we collected (\textsection\ref{sec:data-paired}) to fine-tune our pre-trained models (\textsection\ref{sec:pretrain}) for ASR.
We first outline our modeling approach (\textsection\ref{sec:asr_modeling}) and then scale the number of languages for multilingual ASR from 61 to \nummmslang{} in order to better understand the impact of supporting more languages (\textsection\ref{sec:asr_scaling}).
Next, we compare our models to existing multilingual work (\textsection\ref{sec:asr_main_results}), and build robust multilingual models supporting 1,162 languages trained on several existing corpora as well as the \MMS{} data (\textsection\ref{sec:asr-everything}).
Finally, we evaluate our multilingual models on all languages they support (\textsection\ref{sec:asr-full-eval}).

\subsection{Modeling and Training Approach}
\label{sec:asr_modeling}

We train multilingual speech recognition models by fine-tuning our pre-trained \modelnameb{1} model (\textsection\ref{sec:pretrain}) using labeled data, similar to~\citet{baevski2020wav}. 
To output transcriptions, we add a linear layer on top of the model which maps to an output vocabulary which is the set of letters in the labeled training data of all languages considered in a particular setting.
Next, we fine-tune the entire model with the Connectionist Temporal Classification (CTC) criterion~\citep{graves2006ctc}.

\paragraph{Optimization.}
We use Adam~\citep{kingma2015adam} with exponential decay rates $\beta_1=0.9$, $\beta_2=0.98$ to train model weights using a tri-stage schedule where the learning rate is warmed up for the first 10\% of updates, held constant for the next 40\% updates, and then decayed in the final 50\% updates. 
We experimented with different learning rates ($\Enot{1e-4}$, $\Enot{7e-4}$, $\Enot{3e-4}$ $\Enot{1e-5}$, $\Enot{7e-6}$, $\Enot{3e-6}$, $\Enot{1e-6}$) and number of updates (50K, 100K, 200K, 300K).
Unless otherwise mentioned, we fine-tune models for a total of 50K updates with a batch size of 0.8 hours of data using 16 A100 GPUs with 80GB of memory.

\paragraph{Language-specific Adapters, Head and Fine-Tuning (LSAH).}
In addition to training dense models which share all parameters across languages, we also consider adding adapter modules~\citep{houlsby2019parameter} to models where we use a different set of adapter weights for each language.
Specifically, we introduce adapters at every block of the transformer, and the adapter is added after the last feed-forward block. The adapter module consists of a LayerNorm \cite{ba2016layer} layer, a downward linear projection followed by a ReLU activation and an upward linear projection; the inner dimension of the projections is 16. 

For each language, using an adapter increases the number of parameters by 2M or about 2\% of the total number of parameters without adapters.
We then perform a second stage of fine-tuning for each language where we introduce a randomly initialized linear layer mapping to the specific output vocabulary of a language in addition to language-specific adapter and fine-tune these additional parameters only for another 2K updates on the labeled data of the respective language.

% \paragraph{Hyper-parameters.}

\subsection{Scaling Multilingual ASR to \nummmslang{} Languages}
\label{sec:asr_scaling}

We first analyze training multilingual ASR models with an increasing number of languages by scaling from 61 to \nummmslang{} languages, roughly doubling the number of languages at every step.
Models are trained on the labeled data of \MMS{} by fine-tuning the \modelnameb{1} pre-trained model (\textsection\ref{sec:pretrain}) and we consider training models with and without language specific adapters and output layers.
Each setting is evaluated on the 61 FLEURS languages covered by \MMS{} (FLEURS-61) as well as the 49 languages of CommonVoice covered by \MMS{} (CV-49). 
Results are reported on the development sets of FLEURS and CommonVoice in terms of Character Error Rate without a language model.

\insertResultsASRLangScaling

\autoref{fig:results_asr_scale_lang} shows that performance degrades quickly for dense models which have no language-specific parameters:
for FLEURS-61, CER increases by 5.1 when moving from 61 to \nummmslang{} languages and for CV-49, there is a 2.1 CER increase.
This is mainly due to languages being confused with each other which results in large performance drops for certain languages.
Language-specific parameters (LSAH) alleviate this issue and show only very little degradation (0.4 CER for FLEURS-61 and 0.2 CER for CV-49).

In summary, this shows that scaling multilingual ASR models to over one thousand languages is feasible and that there is little performance degradation when coupled with language-specific parameters.

\subsection{Comparison to Other Work}
\label{sec:asr_main_results}

Next, we present comparisons to other recent related work on multilingual ASR: 
Whisper~\citep{radford2022whisper} uses large quantities of labeled web data to train a model supporting 99 languages (\textsection\ref{sec:asr-comp-whisper}) and Google USM~\citep{zhang2023usm} builds multilingual ASR models supporting 100 languages by pre-training on YouTube data (\textsection\ref{sec:asr-comp-usm}).

\subsubsection{Whisper}
\label{sec:asr-comp-whisper}

Whisper is a multilingual model trained on 680K hours of weakly labeled audio data from the web and able to transcribe speech in 99 languages~\citep{radford2022whisper}.
The model uses a sequence to sequence architecture~\citep{sutskever2014sequence} which has a neural sequence model as decoder that acts in part like a language model.
The decoder has been trained on the target side text of the labeled training data which likely amounts to several billions of words of text from the web.\footnote{Assuming 10K words of text per hour of paired speech data, the decoder was trained on about 6.8B words.}

\insertResultsASRWisper

In contrast, \modelname{} is a CTC model whose decoder is a simple linear layer mapping to a set of characters (\textsection\ref{sec:asr_modeling}).
When comparing CTC models to models with sequence-model based decoders, the former are typically paired with an external language model to enable a fairer comparison.
We therefore train simple n-gram models on web data (Common Crawl) for each language and use it during inference time~(CC LM; \citealt{heafield-2011-kenlm,conneau2020xlsr,nllbteam2022language}); Appendix~\ref{app:train-lms} details the training procedure.
We evaluate both models on the 54 languages of FLEURS supported by both Whisper and \modelname{}  (FLEURS-54) and report word error rate except for  Thai, Lao, Burmese and Khmer where we use character error rate.\footnote{We follow Whisper's evaluation methodology but add Khmer to the list of languages where evaluation is in terms of CER because there is no standard tokenization we are aware of.}

The results (\autoref{tab:asr-whisper}) show that \modelname{} reduces the word error rate of Whisper by a relative 58\% while supporting over 11 times the number of languages.
Moreover, \modelname{} was trained on \nummmshrs{} hours of labeled data compared to 680K for Whisper.\footnote{The non-English portion of the Whisper training data is still 117K hours and covers less than 100 languages, while \MMS{} is less than half the size and covers 11 times the number of languages.}
A reduced version of \modelname{} trained on 61 languages can still outperform Whisper to a similar degree while being trained on only about 3K hours of labeled training data.
Overall, \modelname{} (LSAH) outperforms Whisper on 31 out of the 54 languages.
Appendix~\ref{app:asr-whisper} provides a per-language breakdown of the results.

\subsubsection{Google USM}
\label{sec:asr-comp-usm}

This model is pre-trained on 12M hours of proprietary YouTube audio spanning 300 languages and then fine-tuned to perform ASR for up to 100 languages on a labeled dataset of 90K hours~\citep{zhang2023usm} which results in large improvements over Whisper.\footnote{
We were not able to obtain the list of languages involved in their comparison to Whisper and therefore resort to a comparison not involving the labeled YouTube data. 
This has the downside of removing the impact of the labeled datasets of each approach in the comparison.
}
In a best effort comparison, we adopt the USM setup where the authors fine-tune their pre-trained model on the labeled FLEURS data.
We follow the same training regime as outlined above (\textsection\ref{sec:asr_modeling}) but fine-tune this model for a total of 300K updates.
Results are in terms of character error rate.\footnote{\citet{zhang2023usm} did not perform any additional pre-processing on the reference transcriptions when reporting CER which enables a comparison to their results [Yu Zhang, personal communication 5 May 2023].}

There are several important differences between their approach and \modelname{}:
first, USM uses an RNN-T model~\citep{chan2015las} which has a built-in neural language model while as \modelname{} is a CTC-based acoustic model.\footnote{
RNN-T may benefit particularly from pre-training on unlabeled text data in the case of USM-M, effectively training a strong neural language model.}
Second, some of the USM models were pre-trained on large quantities of unlabeled text as well as 20K hours of labeled audio data (USM-M/USM-M-adapter) while as \modelname{} is pre-trained only on unlabeled speech data.

\insertResultsASRUSM

To enable a fairer comparison, we use n-gram language models trained on unlabeled text during inference: 
for each language, we use either an n-gram model trained on CommonCrawl or a model trained on the transcriptions of the FLEURS training set, depending on dev set performance and data availability.
Appendix~\ref{app:train-lms} details the training procedure.

\autoref{tab:asr-usm} shows that \modelname{} performs very competitively compared to USM.
We note that the approaches have significant differences in the model architecture and uses of unlabeled/labeled data, however, we believe that the result convinces the reader that a simple CTC model paired with n-gram models can perform very competitively to more advanced architectures and more elaborate pre-training procedures.
Appendix~\ref{app:finetune-fleurs-results} shows a per-language breakdown of the results.

\subsection{Robust Multilingual ASR Models}
\label{sec:asr-everything}

\insertTableASREverything

In this section, we turn to building multilingual ASR models on data from multiple domains following a similar approach as prior work for English-only models~\citep{likhomanenko2020rasr,chan2021speechstew,hsu2021robust}. 
We fine-tune the pre-trained \modelnameb{1} model on \MMS{}, FLEURS, CommonVoice, VoxPopuli, and MLS data to support 1,162 language and perform 300K updates; we denote this as multi-domain training.
During fine-tuninig, the data of each language and dataset is balanced similar to pre-training (\textsection\ref{sec:pretrain-setup}; using $\beta_D=0$ and $\beta_L=0.3$). 
The validation set for this model is the concatenation of the dev sets of each dataset for every language.

We evaluate this single model on FLEURS, CommonVoice, VoxPopuli and MLS.
For FLEURS we measure average CER on the 102 languages of FLEURS, for CommonVoice WER over 76 languages, for VoxPopuli WER over 14 languages and for MLS WER over eight languages.
We also fine-tune \modelnameb{1} on each benchmark individually (single-domain training).\footnote{
Models trained on FLEURS data were trained for 300K updates, for MLS models we found 50k updates to work well, CommonVoice and VoxPopuli models were trained for 200K updates.
}
During inference, we use n-gram models trained on CommonCrawl data.

\autoref{tab:asr-everything} shows that the multi-domain model (\MMS{}+FL+CV+VP+MLS) can perform very competitively in several settings:
For both FLEURS and CommonVoice it outperforms prior work as well the single-domain baselines and for VoxPopuli and MLS it is slightly worse than the single-domain baselines.
For VoxPopuli, the \modelname{} model it is outperformed by Maestro which supports a much smaller number of languages and on MLS other approaches are better which attribute to a focus on fewer languages.
Whisper results are not strictly comparable due to the different normalization but it appears to perform better on MLS due to a focus on head languages.
Overall, this demonstrates that a combined model supporting well over 1,000 languages can perform in aggregate competitively on a range of benchmarks.

\subsection{Evaluation on \nummmslang{} Languages}
\label{sec:asr-full-eval}

\insertTableResultsASRFullEval

Finally, we evaluate the multi-domain model trained on \MMS{}, FLEURS, CommonVoice, VoxPopuli and MLS on the test sets of all \nummmslang{} languages in \MMS{}.
We measure character error rate and group languages into six geographical regions covered by \MMS{}: Asia,
North America, South America, Europe, Africa and the Pacific region.
In order to get a broader sense of quality, we measure the number of languages for which CER $\leq$ 5. 
This indicates for how many languages the model makes on average no more than one error in twenty characters.
While this measure is very coarse, it enables us to get a sense of quality across such a large number of languages.

\autoref{tab:asr-full-eval} shows that our model meets the CER quality threshold for 96\% of the \nummmslang{} languages.
The region with the lowest rate is Africa at 91\% and we attribute this in part to different writing scripts. 
We note that this metric is by far not perfect as it imposes the same threshold for every language which may not be appropriate due to different character sets etc.
Also, many of the recordings in \MMS{} are single speaker which means that both the training data and the test data contains utterances with the same voice for a particular language.
While this makes evaluation challenging, we hope that this analysis gives a sense that the model can be used to transcribe a wide variety of languages.

\section{Language Identification}
\label{sec:lid}

Language Identification (LID) is the task of determining the language which is spoken in a given utterance. 
This has several important applications: 
despite much work on multilingual speech recognition~\citep{burget2010multilingual, lin2009study,heigold2013multilingual,bourlard2011current,cho2018multilingual,toshniwal2018multilingual,kannan2019mult,li2019bytes,pratap2020massively}, many deployed systems are still trained on data in a single language despite the need to  transcribe speech in different languages. 
It is therefore crucial to route utterances to the correct system and this routing depends on an LID model.
Moreover, much work on mining speech data from the web relies on LID, including some of the most recent large-scale weakly-supervised work~\citep{radford2022whisper}.
Training a language identification system requires speech data for which the spoken language is known, however, the most diverse public corpora span no more than about 100 languages~\citep{valk2020voxlingua,conneau2022fleurs}.

In this section, we first describe our methodology to build LID models (\textsection\ref{sec:lid-setup}), then evaluate the feasibility of training LID systems using \MMSU{} and \GRN{} data compared data from existing corpora (\textsection\ref{sec:lid-existing-data}) and then build LID models with 40x more languages compared to existing systems (\textsection\ref{sec:lid-scaling}).\footnote{Note that we use \MMSU{} instead of \MMS{} because it supports more languages and we do not require the data to be paired with transcriptions for LID (\textsection\ref{sec:data-paired-source}).}

\subsection{Training Setup}
\label{sec:lid-setup}

We train models by fine-tuning the \modelnameb{1} pre-trained model (\textsection\ref{sec:pretrain-setup}) for language identification.
This is done by stacking a linear classifier on top of the pre-trained model, which maps to the set of possible languages for a particular task, followed by fine-tuning all parameters, including the pre-trained model.

We optimize models with Adam with exponential decay rates $\beta_1=0.9$ and $\beta_2=0.98$ and a tri-state learning rate schedule where the learning rate is warmed up for the first 10\% of updates, held constant for the next 40\% and then linearly decayed for the remainder of training~\citep{kingma2015adam}.
During development, we experiment with different hyper-parameters and perform final model selection based on development set accuracy.
We experiment with different learning rates ($\Enot{1e-5}$, $\Enot{3e-5}$, $\Enot{3e-6}$, $\Enot{5e-6}$, $\Enot{7e-6}$), training updates (10K, 20K, 30K, 40K, 50K) and batch sizes (1.5min, 3min, 6min). 
We train models on 16 GPUs.

To balance the different languages and corpora during training, we first balance the data of every language in the different corpora, using sampling parameter $\beta_L$. This is followed by balancing each language using the resampled data of the first step, using sampling parameter $\beta_D$ with the sampling distribution outlined in~\textsection\ref{sec:pretrain-w2v}.
We experiment with $\beta_L$ and $\beta_D$ settings 0, 0.3, 0.5, 0.7 and 1.

When we train models on multiple datasets, then we use a development set containing up to 30 minutes of data for each language and we sample an equal amount of data from each corpus.
For models supporting 1K, 2K or 4K languages, we reduce the amount of data per language to 15 min, 7min and 3min to enable faster development.

\subsection{Comparison to Existing Datasets}
\label{sec:lid-existing-data}

\insertResultsLIDFLVLCompare

We first assess the effectiveness of training LID models on \MMSU{} and \GRN{} data compared to existing LID training data (FLEURS, VoxLingua-107).
Performance is evaluated both on the FLEURS and VoxLingua-107 benchmarks.
In this setting, FLEURS and VoxLingua-107 models are at an advantage because there is no domain shift compared to systems trained on \MMSU{} and \GRN{} data.

We are particularly interested in the setting where models trained on existing corpora are evaluated out-of-domain, i.e., the performance of a model trained on VoxLingua-107 evaluated on FLEURS or a model trained on FLEURS evaluated on VoxLingua-107 data, and how this compares to models trained on \MMSU{}/\GRN{}.
To enable a controlled comparison, we report LID accuracy on the 72 languages supported by all considered datasets and train only on data of these languages.

The results (\autoref{fig:lid-fl-vl-comp2}) show that \MMSUboth{} enables LID with slightly lower performance compared to systems trained on existing datasets when both are evaluated out-of-domain: 
on FLEURS evaluation data the gap is 2.1\% compared to a VoxLingua-107 model and on VoxLingua-107 evaluation data \MMSUboth{} trails a FLEURS model by 1.6\%.
Combining \MMSU{} and \GRN{} works particularly well as \GRN{} is more varied which improves performance.
Naturally, models trained on in-domain training data (FLEURS or VoxLingua-107) perform best.
We conclude that \MMSU{} and \GRN{} enable good quality LID models while enabling LID for many more languages as we will demonstrate next.

\subsection{Scaling Language Identification to \numlidlang{} Languages}
\label{sec:lid-scaling}

\insertLIDResultsScaling

Next, we scale spoken language identification from about 100 languages to \numlidlang{} languages by combining \MMSU{}, \GRN{}, FLEURS, and VoxLingua-107 data.
Our primary goal is to understand how accuracy is impacted as models support more and more languages.
We start with 126 languages, the union of the languages in both FLEURS and VoxLingua-107, and then add languages from \MMSUboth{}, roughly doubling the number of languages at every experiment.
Languages are sorted by descending speaker count and we add the most spoken languages first.\footnote{We obtain the number of speakers of each language from \url{https://www.ethnologue.com}}

Performance is evaluated on FLEURS and VoxLingua-107 which is in-domain with respect to the models we train.
To get a sense of how the models perform on domains not seen in the training data, we also evaluate on two other datasets which are out-of-domain with respect to the training data domain:
first, BABEL is conversational telephone speech data and we evaluate on 23 African and Asian languages~\citep{gales2014babel}.\footnote{Amharic, Assamese, Bengali, Cantonese, Cebuano, Georgian, Guarani, Haitian Creole, Igbo, Javanese, Kazakh, Lao, Lithuanian, Dholuo, Mongolian, Pashto, Swahili, Tagalog, Tamil, Telugu, Turkish, Vietnamese,  Zulu. 
We evaluate on test utterances longer than 10 seconds.}
Second, VoxPopuli~\citep{wang2021voxpopuli} consists of parliamentary speech in 25 languages from the European parliament. 
We evaluate on 2.5 hours of data for each language sampled from the VoxPopuli unlabeled data portion.

\autoref{tab:lid-scale} shows that \modelname{} models scale very well: 
increasing the number of languages from 126 to \numlidlang{} results in a modest performance drop of just 0.3\% on FLEURS and no drop on VoxLingua-107.
Out-of-domain we observe a drop of 3.6\% on BABEL and 0.2\% on VoxPopuli.
The results are also competitive to models trained only on in-domain data: 
On FLEURS evaluation data, the \numlidlang{} language \modelname{} model performs 1\% better than the baseline trained only on FLEURS data and is only 0.2\% behind the baseline trained on both FLEURS and VoxLingua-107.
On VoxLingua-107 evaluation data, the gap is 0.4-0.8\%.
This shows that scaling LID to \numlidlang{} can result in models with competitive performance to models trained on much fewer languages.

\section{Speech Synthesis}
\label{sec:tts}

As a final downstream task we consider speech synthesis or text-to-speech (TTS) where models output speech for a corresponding input text.
Most prior work focuses on English using clean corpora and high quality phonemizers~\citep{tan2021survey}.
These resources are only available for a small number of languages which is why expanding TTS to \nummmslang{} languages requires different choices.
\citet{black2019cmu} built TTS systems for 699 languages using data from a similar source as \MMS{}.
Encouraged by our initial results indicating higher quality data (\textsection\ref{sec:data-comp-cmu}), we extend their work by building models for \nummmslang{} languages.

We first describe the model architecture on which we build (\textsection\ref{sec:tts-model}), 
how we pre-process the data to train TTS models (\textsection\ref{sec:tts-preprocess}) and how we evaluate our models (\textsection\ref{sec:tts-eval}).
Next, we present an ablation of our design choices compared to a highly optimized setup used for English (\textsection\ref{sec:tts-results-english}).
We also measure performance when synthesizing out-of-domain data for 61 languages of the FLEURS benchmark (\textsection\ref{sec:tts-results-fleurs}), and finally present results for all \nummmslang{} languages (\textsection\ref{sec:tts-results-all}).

\subsection{Text-To-Speech Model}
\label{sec:tts-model}

Our Text-to-Speech (TTS) model is based on VITS~\citep{kim2021conditional}, which is one of the state-of-the-art TTS approaches. 
While VITS has previously been applied in a multilingual setting for English, Portuguese, and French~\citep{casanova2022yourtts}, we scale it to 1,107 languages.

VITS is an end-to-end speech generation network that predicts the raw speech waveform from a text sequence. 
It can be viewed as a conditional variational auto-encoder (VAE;~\citealt{Kingma2013AutoEncodingVB}) that estimates audio features based on the input text. The spectrogram-based acoustic features are generated by a flow-based sub-network, which includes multiple affine coupling layers~\citep{dinh2017density} and a text encoder. 
The waveform is decoded using a stack of transposed convolutional layers that have been adapted from HiFi-GAN~\citep{kong2020hifi}. The model is trained end-to-end with a combination of losses derived from variational lower bound and adversarial training. During inference, the text encodings are upsampled based on an internal duration prediction module and then mapped into the waveform using a cascade of the flow module and HiFi-GAN decoder.

We train separate VITS models for each language. 
Most of our hyperparameters are identical to the VITS model trained on LJSpeech~\citep{ljspeech17,kim2021conditional} except for these differences:
Instead of training the model for 800K steps, we train each model for 100K steps using eight V100-GPUs with a batch size of 64 per GPU.
This setup was only slightly worse than the original configuration but reduced training time by approximately eight times which made training a large number of TTS systems feasible (\textsection\ref{sec:tts-eval}).
We experimented with different learning rate settings and found that the original learning rate schedule of VITS worked best.

\subsection{Text and Speech Data Pre-processing}
\label{sec:tts-preprocess}

\paragraph{Data Selection.}
To train models we use the \MMS{} dataset which provides paired speech and text data.
For most languages we have a single recording of the New Testament, however, for 99 languages multiple recordings are available (\textsection\ref{sec:data-paired-source}). 
For these languages, we choose a single recording in order to avoid introducing additional speakers into the training data.
To choose a recording, we train ASR models on the data and select the recording based on which the corresponding ASR model achieves the lowest CER on a held-out set of an out-of-domain evaluation set. 
If no out-of-domain evaluation set is available, we choose a random recording. 
Finally, if there are both drama and non-drama recordings, then we consider only non-drama recordings.

\paragraph{Text Representations.}
Most current TTS models convert text to phonemes using grapheme-to-phoneme tools such as g2p~\citep{g2pE2019} which rely on a lexicon to map input characters to phonetic representations.
However, such lexicons are not available for most low resource languages since they require manual annotation.
In order to scale TTS to over one thousand languages, we represent the input text as individual letters for languages with a small vocabulary.
For languages with 200 or more characters, we use a uroman encoding (\citealt{hermjakob-etal-2018-box}; \textsection\ref{sec:data-initial-align}).
We validated this strategy on the languages of FLEURS where found that letter-based models outperform uroman-based systems across all languages, except for Amharic and Korean, which both have large character sets of between 200-1,000 characters.
We therefore used this mixed strategy.\footnote{
For all \nummmslang{} languages, the following languages use a uroman encoding: Amharic, Gumuz, Korean, Sebat Bet Gurage and Tigrinya.}

\paragraph{Speech Data Pre-processing.}
For drama recordings, we remove background music to enhance the quality of TTS models.
We use a denoiser model~\citep{defossez2020real} to remove background music.
We also noticed that some utterances contain multiple speakers, usually voicing different characters in the read stories. 
We use a simple heuristic to detect those utterances and remove them from the training data.
The heuristic computes the variance of the pitch on voiced frames and removes utterances with high variance in the pitch; the pitch estimation is based on~\citet{Mauch2014PYINAF}.
We found that removing 15\% of the utterances with the highest pitch variance in each recording removed many multi-speaker utterances.

\subsection{Evaluation Methodology}
\label{sec:tts-eval}

Evaluation of speech synthesis is not straightforward even in the most researched English setting comprising a clean corpus with single speaker data~\citep{ljspeech17} and expanding evaluation to a large number of languages poses additional challenges.
Similar to prior work, we rely on both automatic metrics (MCD and ASR) and human studies which we detail next.

\paragraph{Mel-Cepstral Distortion (MCD).}
Mel-cepstral distortion is an automatic metric which measures the closeness of synthesized speech to a human utterance for the same text in terms of the warping distance of mel frequency cepstral coefficients.
There are two disadvantages of MCD:
first, it is only meaningful when the voice and prosody of the speaker as well as recording conditions in the training data matches the evaluation data because the mel cepstral sequences are sensitive to these aspects.
This prevents evaluation on data outside the \MMS{} domain.
Second, it is not well suited to measuring the intelligibility of the TTS output.
We address the former by focusing MCD evaluation on \MMS{} data and the latter via the next metric.

\paragraph{Automatic speech recognition (ASR).}
Transcribing the synthesized speech with automatic speech recognition and then measuring the error rate has the potential to address both issues of MCD: it can be measured on evaluation sets which are out-of-domain and it captures the content of the synthesized speech.
Specifically, we synthesize both in-domain data from the \MMS{} evaluation sets as well as out-of-domain data from the FLEURS corpus. 
Next, we apply ASR models and compute CER of the ASR model output with respect to the input text of the TTS system.
Low CER indicates that the TTS model is able to capture the content of the input text.
Unless otherwise stated, we use ASR models trained on FLEURS data.

\paragraph{Mean Opinion Score (MOS).}
Finally, we evaluate models using MOS and because it is very hard to find human raters proficient for a large number of languages, we ask raters to judge the fidelity of the synthesized samples together with how natural the speech sounds.
We rely on ASR to get a sense of how much of the content is preserved in the TTS outputs and consider the MOS score for generation quality and naturalness.
We evaluate 50 samples per method of each language, collect ten judgements per sample and ask raters to judge the quality from 1 (lowest quality) to 5 (highest quality). 
Results are reported in terms of the mean score as well as the 95\% confidence interval.
We use the CrowdMOS~\citep{ribeiro2011crowdmos} package with the recommended recipes for detecting and discarding inaccurate ratings.
We do not require raters to be able to speak the respective language (except for English) as it is very hard to find raters that speak all the languages we consider. 
We rely on the ASR error rate to get a sense of content preservation in the TTS models.

\subsection{Evaluation of Design Choices}
\label{sec:tts-results-english}

\subsubsection{Training Setup}

We first analyze the quality of our training setup (\textsection\ref{sec:tts-model}) and compare it to the common VITS setup~\citep{kim2021conditional} which trains models for up to 800K updates on the clean LJSpeech corpus~\citep{ljspeech17} with text data pre-processed by a high-quality phonemizer~\citep{g2pE2019}. 
This contrasts to our setup where we train the same model for 100K updates on the \MMS{} data using letters.

\insertTableResultsTTSDesign
\insertTableResultsTTSDrama

We evaluate performance on the development sets of \MMS{} which is in-domain for \MMS{} models, LJSpeech (LJS; \citealt{ljspeech17}) which is in-domain for the original VITS setup and FLEURS which is out-of-domain for both.\footnote{For measuring CER on LJSpeech, we remove punctuation and capitalization of both references and hypothesis using the normalization of~\citet{radford2022whisper}.}
The audio of FLEURS frequently contains high levels of noise and reverberation and it is relatively uncommon to use it for TTS, however, it does cover a large number of languages and it is out-of-domain with respect to the \MMS{} training data which makes it an interesting evaluation set, particularly for the experiment in \textsection\ref{sec:tts-results-fleurs}.

\autoref{tab:tts-design} shows that our reduced setup (row 5) generally performs less well than the highly optimized setup of VITS for LJSpeech~(row 2; \citealt{kim2021conditional}) and each design choice (fewer training updates, \MMS{} training data and character inputs) leads to a reduction in quality.
In terms of character error rate, the degradation is most pronounced on out-of-domain settings with respect to our reduced setup (row 5) and less so on the in-dmain \MMS{} development set.
We stress that these choices enable scaling TTS to over 1,000 languages at manageable compute requirements and no need for language-specific text processing tools.

Note that the CER of almost all models on FLEURS data is lower than for the corresponding natural speech and that the MOS scores of the human reference utterances is also lower than for other datasets.
We attribute this to the high levels of noise and reverberation in the FLEURS audio which results in increased CER for the original samples compared to the CER of the synthesized samples.

\subsubsection{Data with Background Music}

Next, we ablate how we build speech synthesis models based on recordings with background music which is a setting that applies to about 38\% of the languages (\textsection\ref{sec:data-paired-source}).
For this purpose, we train a model on an English recording with background music before and after the pre-processing steps outlined in~\textsection\ref{sec:tts-preprocess} and compare this to a model trained on another English recording that does not contain any background music to start with (no background music).
The pre-processing denoises the data and removes utterances with multiple speakers.

The results (\autoref{tab:tts-drama}) show that both denoising and removing samples with multi-speaker utterances results in performance improvements compared to the original data with background music.
Both steps reduce the CER gap to models trained on no background music data by 69-87\% relative to the CER of the system trained on data with background music.
MOS scores also increase after pre-processing, sometimes close to the level of the data with no background music on \MMS{} evaluation data (3.47 vs. 3.51).

\subsection{Out-of-Domain Evaluation}
\label{sec:tts-results-fleurs}

The \MMS{} data is from a particular narrow domain (\textsection\ref{sec:data-paired-source}) which poses the question of whether TTS models trained on this data will generalize to other domains.
To get a better sense of this, we train speech synthesis models and evaluate their quality on both in-domain and out-of-domain data.
As in-domain data we use the test sets of \MMS{} and as out-of-domain data we use FLEURS. 
This enables evaluation on 61 languages of the FLEURS benchmark (FLEURS-61) which are covered by \MMS{} data. 
However, a downside of FLEURS is that it contains high levels of noise which makes it challenging when comparing synthesized audio to the human reference audio.

\insertTableResultsTTSFLEURS

\autoref{tab:tts-fleurs} shows that \modelname{} models are robust to domain shift:
the CER of synthesized speech (TTS) is only slightly higher out-of-domain compared to in-domain and MOS scores for the synthesized samples are nearly identical.
We note that the in-domain and out-of-domain settings are measured on different data which does not enable strong claims about identical performance.
The systems also retain much of the original content as the small difference in CER between TTS and human utterances (ref) shows.

The MOS scores also indicate that our systems have lower sound quality compared to human utterances but the difference is not very large on in-domain data (3.51 vs. 3.61).
Unfortunately, out-of-domain MOS scores for the references are affected by the noisy speech in the FLEURS audio as noted earlier.
We conclude that TTS models trained on \MMS{} data perform well out-of-domain.

\subsection{Evaluation on \nummmslang{} Languages}
\label{sec:tts-results-all}

Finally, we train models for all languages of \MMS{} and focus on in-domain evaluation since finding out-of-domain evaluation data for such a large number of languages is difficult.
Specifically, we measure MCD and ASR CER on the \MMS{} test sets. 
To be able to evaluate ASR quality on all languages, we use ASR models trained on \MMS{} data which results in much lower error rates.
Similar to \textsection\ref{sec:asr-full-eval}, we group results into six geographical regions covered by \MMS{}: Asia, North America, South America, Europe, Africa and the Pacific region.

To get an overall sense of how many models are of good quality, we measure whether the model for a particular language has ASR CER $\leq$ 5. 
This indicates the number of systems which make on average no more than one error in twenty characters.
While this measure is by far not perfect, it enables us to get a broad sense of quality across a large number of languages.

\insertTableResultsTTSFull

\autoref{tab:tts-full} shows that about 85\% of the \nummmslang{} languages meet the CER quality threshold. 
South American and European languages achieve the highest rate at 95\% and African languages the lowest rate of 76\%. 
This is in part driven by different writing scripts.
The ASR character error rates are generally low because the error rates are based on ASR models trained on \MMS{} data.

\section{Bias Analysis and Ethical Considerations}

Training machine learning models on religious texts may introduce biases and requires ethical considerations. 
In this section, we analyze whether our models are biased to perform better for different genders (\textsection\ref{sec:rai-gender-bias}), if the language produced by our models is religiously biased (\textsection\ref{sec:rai-lang-bias} and finally, we discuss ethical considerations of using religious texts in research (\textsection\ref{sec:rai-ethical}).

\subsection{Understanding Gender Bias}
\label{sec:rai-gender-bias}

Most speakers in \MMS{} dataset appear to be male and this bears the risk of machine learning models trained on this data performing better for male speakers.
To understand whether the models trained on our datasets (\textsection\ref{sec:datasets}) exhibit gender bias, we perform the following experiment:
we evaluate models on the development set of FLEURS which contains metadata indicating the gender of the speaker and we use this information to report performance for each gender.
Using this split, we evaluate the accuracy of ASR models trained on \MMS{} on 27 languages of the FLEURS dataset for which \MMS{} provides data (61 languages), and for which there are at least 50 samples for each gender (27 out of these 61 languages).

\insertGenderAnalysis

\autoref{fig:rai-gender-analysis} shows that the average character error rate over these 27 languages is very similar, both for the \modelname{} model and the model trained on FLEURS data.
There can be significant differences between genders within a particular language, but both models appear to be equally affected (see~\autoref{tab:rai-gender-full-results}).
On a per language basis, male speakers have a higher error rate for 14 languages while as female speakers have a higher error rate for the remaining 13 languages.
We conclude that our models exhibit similar gender bias to models trained on FLEURS data which is general domain data.

\subsection{Understanding Language Bias}
\label{sec:rai-lang-bias}

The datasets created as part of this study are from a particular narrow domain and machine learning models estimated on this data may exhibit certain biases.
In this section, we examine the extent to which automatic speech recognition models trained on \MMS{} data output biased language.

\paragraph{Methodology.}
The general methodology of our analysis is to identify a set of biased words in each language and to measure the rate at which biased words are produced by ASR models.
We compare models trained on \MMS{} data and models trained on FLEURS, a corpus of people reading Wikipedia articles in different languages.
Both sets of models are tasked to transcribe the FLEURS development set and we are interested in whether models trained on \MMS{} data are more likely to use biased language compared to models trained on FLEURS data.

\paragraph{Identifying Biased Words.} 
We were not able to find speakers for most of the considered languages of this study and therefore use the following automatic procedure to determine religious words: 
for each word that occurs in the training data of \MMS{}, we compare the relative token frequency, that is, the rate at which the word type occurs in the \MMS{} data, to the relative token frequency in a general domain corpus; we use Common Crawl~\citep{conneau2020xlsr} as a general domain corpus.
If the relative word frequency is at least twice as high in \MMS{} compared to Common Crawl, then we add it to the subset of words we include in our study.
This enables us to evaluate on 51 languages of the FLEURS corpus since not all languages are covered by \MMS{} and we also need to find data in Common Crawl for each language.
The automatic procedure has the added benefit of avoiding any potential biases introduced by human annotators.

\paragraph{Results.}
\autoref{fig:data-rai-words} shows that the rate of biased words is much lower in the outputs of \modelname{} models compared to the training data (\MMS{} ASR output vs. \MMS{} train).
For many languages, \modelname{} models generate these words at the same rate as the FLEURS models.
On average the rate of biased words is 0.7\% absolute higher for \modelname{} compared to the FLEURS model.
We interpret this as a slight bias as the difference between \MMS{} ASR output and FLEURS ASR output in~\autoref{fig:data-rai-words} shows.

We consulted with native speakers for languages which showed the largest discrepancies:
for Mongolian (mon), a native speaker verified that most of the biased words in question are actually general language with no particular bias.
For Persian (fas), only two out of the words the procedure identified were of religious nature and both were indeed over predicted: the Persian words for Jesus and spirit/ghost were predicted in four instances (on the entire development set) while the FLEURS model does not predict these words, however, the \modelname{} model also predicts the word for hand 15 times more often than the baseline FLEURS model.

In English, the procedure identifies words such as \emph{you}, \emph{that}, \emph{they} but it also includes \emph{jesus}, which both models predict at the same rate. There is also \emph{christ}, and \emph{lord}, both of which are not predicted at all, or \emph{men} which is predicted six times by the \modelname{} model compared to five times by the FLEURS model.
Our method of identifying biased words has low precision but it does capture words which are likely more used in religious contexts than otherwise.

\insertDataRAIWords

\subsection{Ethical Considerations and use of Religious Texts in Research}
\label{sec:rai-ethical}

Our consultations with Christian ethicists concluded that most Christians would not regard the New Testament, and translations thereof, as too sacred to be used in machine learning.
The same is not true for all religious texts: for example, the Quran was originally not supposed to be  translated.
There is also the risk of religious training data biasing the models with respect to a particular world view, however, our analysis of the language generated by our models suggests that the language produced by the resulting speech recognition models exhibit only little bias compared to baseline models trained on other domains (\textsection\ref{sec:rai-lang-bias}).

This project follows a long line of research utilizing the New Testament to train and evaluate machine learning models.
The most related project is the CMU Wilderness effort~\citep{black2019cmu} which created speech synthesis models for 699 languages using speech and text data from similar sources as our datasets.
For machine translation, researchers used data from the Bible both for training and evaluation~\citep{christos2014bible,mccarthy2020bible,nllbteam2022language}.
For speech processing, researchers trained speech synthesis models for ten African languages based on readings of the bible~\citep{meyer2022bibletts}.

\section{Conclusions and Open Problems}

We presented the first study which scaled speech technology to over one thousand languages. 
This has been made possible by the rapid progress in self-supervised speech representation learning which in turn enabled more sample efficient learning from labeled data. 
We presented how we collected datasets, pretrained models and then built models for automatic speech recognition, language identification and speech synthesis.
This scaled the number of supported languages for several major speech tasks by between 10-40x. Going forward, we see several avenues for future work:

\paragraph{Scaling to even more languages and dialects.}
Even though, we built speech systems supporting between 1,100-4,000 languages, there are currently over 7,000 languages being spoken around the world today. 
Moreover, there are many more dialects which are often not adequately represented in the training data, even for high-resource languages such as English. 
This can lead to undesirable biases in the performance of these models~\citep{koenecke2020racial}.

\paragraph{Multi-task models.}
Another avenue is to train single models for several downstream tasks such as speech recognition, language identification etc. which can then all be performed by a single model. 
There has been work on a moderate number of languages~\citep{radford2022whisper} but we hope that this approach can be scaled to many more languages and with a smaller focus on head languages.

\paragraph{Tackling more speech tasks.}
While this study covered three different speech tasks, there are many more tasks involving speech data, such as speech translation, both to text and to speech, or keyword spotting, intent classification etc. We hope that future work will expand these tasks to many more languages as well.

\section*{Acknowledgments}

We would like to thank Chlo\'{e} Bakalar and Hubert Etienne for providing feedback on responsible AI evaluations, as well as the people who featured in the demo video: 
Filmawit Gebreegziabher, Gilbert Buada, Shanti Jha, Masoud Tavazoei, Akinniyi Akinyemi, Mari Silvia Marin, Amlan Roy, Kumar Rajarshi, Sugat Nayak, Jason Carlo Carranceja, Amit Prakash, Hangameh Vahabzadeh, Mohsen Dalil, Swati Verma, Edwina Anandita, Faisal Putra.
We would like to thank Moto Hira for help in preparing the forced alignment tutorial in TorchAudio and Ping Yu for analyzing the Chinese language datasets for \modelname{}.
We thank Sainbayar Sukhbaatar for his analysis of the model transcriptions.

\bibliography{refs}

\bibliographystyle{abbrvnat}

\newpage
\appendix

\appendix
\setcounter{table}{0}
\setcounter{figure}{0}
\renewcommand{\thetable}{A\arabic{table}}
\renewcommand{\thefigure}{A\arabic{figure}}

\section*{Appendices}

\section{Forced Alignment}
\label{app:align}

Given an input audio sequence $X = (x_1,...,x_N)$, where $N$ is the number of frames, a CTC alignment model produces an output sequence $Y = (\bf{y^1,...,y^{T}})$ of length $T$, where $\bf{y^t}$ denotes the posterior probability distribution over an alphabet $\Alphabet$ and blank token $\Blank$. 
Let $\CollapseFunc$ denote the collapsing function of CTC which collapses all repeating symbols and which removes all blanks for a given sequence. 
An alignment path $\boldsymbol{\pi} = \pi_1,...,\pi_T$ in CTC for a given target label of length $M$, $L = (l_1,...,l_M)$ where $\pi_t \in \Alphabet \cup \{\Blank\}$ and $l_i \in \Alphabet$ should satisfy $\CollapseFunc(\boldsymbol{\pi}) = L$. 

Among the all the alignment paths which can be collapsed to the given target label $L$, forced alignment computes the best alignment path $\hat{\boldsymbol{\pi}}$ which maximizes the probability under the posterior distribution given by the acoustic model. 

\begin{equation} \label{eqn:force_align}
\hat{\boldsymbol{\pi}} = \argmax_{\boldsymbol{\pi} \in \CollapseFunc^{-1}(L)} P(\boldsymbol{\pi} | X) 
\end{equation}

CTC assumes every output is conditionally independent of the other outputs given the input. We have 
\begin{equation}
    P(\boldsymbol{\pi} | X) = \prod_{t=1}^{T} P(\boldsymbol{\pi_t} | t, X) = \prod_{t=1}^{T} y^t_{\pi_t}
\end{equation}

\autoref{eqn:force_align} can be computed efficiently using dynamic programming and backtracking.  
An efficient version of this algorithm is implemented in flashlight~\citep{kahn2022flashlight} on CPU. 
We implement a parallel version on CUDA, incorporating memory optimizations that offload memory to CPU, allowing it to process long audio files efficiently as shown in Algorithm~\ref{algo:force_align}.

\begin{algorithm} 
\caption{Pseudo code of our CTC Forced Alignment algorithm on GPU}
\renewcommand{\algorithmicrequire}{\textbf{Input:}}
\renewcommand{\algorithmicensure}{\textbf{Output:}}
\algblockdefx{FORALLP}{ENDFAP}[1]%
  {\textbf{for all }#1 \textbf{do in parallel}}%
  {\textbf{end for}}
\begin{algorithmic}[1]
    \Require{Posterior probabilities $\bf{y}$, target label $L$}
    \Ensure{Alignment path $\pi$} \\ 
    $ S \leftarrow 2 \times |L| + 1$  \\ 
    Create GPU matrices $\alpha_{\text{odd}}$, $\alpha_{\text{even}}$ of size $S$ each to store the maximum log probability of aligning the target upto the given node \\
    B $\leftarrow$ 100; \# buffer size for data copy from GPU to CPU \\ 
    Create CPU matrix $\beta$ of size $T \times S$ for saving the label indices from previous time step used to compute $\alpha_{\text{odd}}$/$\alpha_{\text{even}}$ \\ 
    Create GPU matrix $\beta_{\text{buffer}}$ of size $B \times S$, where B is the buffer size, to store the values of $\beta$ temporarily in a buffer before being copied to CPU. 
    % \zni{should it be updated to (min(B, T), S)?} \vin{This is an implementation detail. The algo works with B x S as well.}
    \For {$t=1,\ldots,T$}
        \FORALLP {$l=1,\ldots,S$} 
            \State If $t$ is odd, update $\alpha_{\text{odd}}[l]$ based on $\alpha_{\text{even}}$, $\bf{y^t}$ using dynamic programming and vice-versa
            \State Store the index $l'$ from previous time step used to update $\alpha_{\text{odd}}$/$\alpha_{\text{even}}$ in  $\beta_{\text{buffer}}$
        \ENDFAP
        \If {t \% B == 0 \textbf{or} t == T}
            \State Copy $\beta_{\text{buffer}}$ to CPU matrix $\beta$ asynchronously
        \EndIf 
    \EndFor
    \\Backtrack and compute $\pi$ from $\alpha_{\text{odd}}$, $\alpha_{\text{even}}$ and $\beta$
\end{algorithmic} 
\label{algo:force_align}
\end{algorithm}

\paragraph{Relationship to Other Alignment Generation Approaches.}
For dealing with noisy transcripts, a commonly used alternative to forced alignment is as follows~\citep{povey2011kaldi}:
first segment the audio into shorter segments that can be input to an acoustic model composed with a language model and generate a transcription for each segment.\footnote{\url{https://github.com/mozilla/DSAlign}} 
Then align the generated transcription with the original text to determine the audio segment for each word. 
If the generated transcriptions cannot be well aligned, then these segments are discarded.

The advantage of this approach is that it enables alignment when the audio and the text do not correspond entirely to each other. 
However, the alignments are performed on a segment per segment basis, whereas our forced alignment process performs a global alignment taking the entire sequence into account.

% \clearpage
\section{\textit{n}-gram Language Models} 
\label{app:train-lms}

We train 5-gram language models on Common Crawl data using KenLM~\citep{heafield-2011-kenlm} for each language in FLEURS.
For languages that do not use spaces to separate words, we train 20-gram character-level language models.
These languages are Mandarin Chinese (cmn), Cantonese Chinese (yue), Japanese (jpn), Thai (tha), Lao (lao), Burmese (mya) and Khmer (khm).
The text is pre-processed following~\autoref{sec:data-preprocess} and we also remove emojis.\footnote{\url{https://stackoverflow.com/a/33417311}}.

For word-level models, we limit the training data to 40GB and select the 250K most-frequent words as the vocabulary. 
For character-level models, we limit the training data to 6GB.
We provide example commands used for training the LMs below:

\shellcmd{\# word-level LM}\\
\shellcmd{> kenlm/build/bin/lmplz  --prune 1 2 3 4 5 -o 5 --limit\_vocab\_file vocab.txt  -S 90\% -T /tmp/ < input.txt > output.arpa }\footnote{We use \texttt{--prune 1 1 2 3 4} if data size < 5GB and additionally use \texttt{--discount\_fallback} if data size < 1GB }  \\

\shellcmd{\# character-level LM} \\
\shellcmd{> kenlm/build/bin/lmplz  --prune 0 0 0 0 0 1 1 1 2 3 -o 20 trie   -S 90\% -T /tmp/ < input.txt > output.arpa} \\

For n-gram models trained on the FLEURS training data transcriptions, we build 15-gram character level language models without any pruning on all languages in FLEURS. 
For the comparison with Whisper, we only use the Common Crawl language models. 

We use the CTC beam-search decoder from the Flashlight~\citep{kahn2022flashlight} library for decoding our models. 
For decoding with word-level LMs, we use the lexicon-based decoder of~\citet{collobert2016wav2letter, pratap2019w2l} and for character-level LMs, we use the lexicon-free beam-search decoder of~\citet{likhomanenko2019convlm}. 
We tune the language model weight and word insertion penalty on the validation set to select the best hyperparameters for decoding the test set.

\newpage
\section{Comparison to Whisper}
\label{app:asr-whisper}

\autoref{tab:asr-whisper-perlang-test} shows a breakdown of the results into individual languages and \autoref{tab:asr-whisper-full} shows results with and without CC LM n-gram language models.

\insertResultsASRWisperPerLang

\insertResultsASRWisperFull

\clearpage
\section{Comparison to USM}
\label{app:finetune-fleurs-results}

\insertResultsASRUSMFull

% \clearpage
\section{Gender Bias Study}
\label{app:gender-bias-study}

\insertResultsGenderBiasFull

\end{document}